\documentclass[10pt,journal,compsoc]{IEEEtran}
\ifCLASSOPTIONcompsoc
\usepackage[nocompress]{cite}
\else
\usepackage{cite}
\fi
\ifCLASSINFOpdf
\else
\fi
\usepackage{amsmath}
\usepackage{amssymb}
\usepackage{graphicx}
\usepackage{algorithm}
\usepackage{algorithmic}
\usepackage{multirow}
\usepackage{array}
\usepackage{mathtools}
\usepackage{xcolor}
\usepackage{times}
\usepackage{ragged2e}
\usepackage{mathrsfs}

\usepackage{epsfig}
\usepackage{color}
\usepackage{subfigure}
\usepackage{diagbox}
\usepackage{bbding}
\usepackage{makecell}
\usepackage[defaultcolor=black]{changes}
\usepackage{enumerate}
\usepackage{booktabs}
\usepackage{subfigure}
\usepackage{threeparttable}
\usepackage[square, comma, sort&compress, numbers]{natbib}
\def\bm #1{\boldsymbol{#1}}

\makeatletter
\DeclareRobustCommand\onedot{\futurelet\@let@token\@onedot}
\def\@onedot{.}
\def\eg{\textit{e.g}\onedot} 
\def\ie{\textit{i.e}\onedot} 
 
\def\etc{{etc}\onedot} 
\def\wrt{w.r.t\onedot} 
\def\etal{\textit{et al}\onedot~}

\def\na{N/A}
\definecolor{sgreen}{rgb}{0.2,0.6,0.15}
\definecolor{sblue}{rgb}{0,0.3,0.9}
\makeatother

\newcommand{\Tref}[1]{Tab.~\ref{#1}}

\newcommand{\Fref}[1]{Figure~\ref{#1}}
\newcommand{\Sref}[1]{Section~\ref{#1}}

\newcommand{\eref}[1]{Eq.~(\ref{#1})}
\newcommand{\fref}[1]{Fig.~\ref{#1}}
\newcommand{\sref}[1]{Sec.~\ref{#1}}

\newcommand{\first}[1]{\underline{#1}}
\newcommand{\second}[1]{\underline{#1}}
\newcommand{\third}[1]{\underline{#1}}

\usepackage[colorlinks,linkcolor=blue]{hyperref}

\makeatletter
\newcommand\figcaption{\def\@captype{figure}\caption}
\newcommand\tabcaption{\def\@captype{table}\caption}
\makeatother

\begin{document}
	%
	\title{Appearance-based Gaze Estimation with Deep Learning: A Review and Benchmark}
	%
	%
	%
	%
	
	\author{Yihua Cheng\textsuperscript{\rm 1}, Haofei Wang\textsuperscript{\rm 2}, 	Yiwei Bao\textsuperscript{\rm 1}, Feng Lu\textsuperscript{\rm 1,2,*}
		\IEEEcompsocitemizethanks{\IEEEcompsocthanksitem Feng Lu is the Corresponding Author.\protect\\
		}
		
		\textsuperscript{\rm 1}State Key Laboratory of Virtual Reality Technology and Systems, SCSE, Beihang University, China.\\
		\textsuperscript{\rm 2}Peng Cheng Laboratory, Shenzhen, China.\\
		\{yihua\_c, baoyiwei, lufeng\}@buaa.edu.cn, wanghf@pcl.ac.cn
	}
	
	%
	%

	\markboth{Journal of \LaTeX\ Class Files,~Vol.~14, No.~8, August~2015}%
	{Shell \MakeLowercase{\textit{et al.}}: Bare Demo of IEEEtran.cls for Computer Society Journals}
	%



	\IEEEtitleabstractindextext{%
		\justifying
		\begin{abstract}

            Human gaze provides valuable information on human focus and intentions, making it a crucial area of research. Recently, deep learning has revolutionized appearance-based gaze estimation. However, due to the unique features of gaze estimation research, such as the unfair comparison between 2D gaze positions and 3D gaze vectors and the different pre-processing and post-processing methods, there is a lack of a definitive guideline for developing deep learning-based gaze estimation algorithms.
            In this paper, we present a systematic review of the appearance-based gaze estimation methods using deep learning. Firstly, we survey the existing gaze estimation algorithms along the typical gaze estimation pipeline: deep feature extraction, deep learning model design, personal calibration and platforms. Secondly, to fairly compare the performance of different approaches, we summarize the data pre-processing and post-processing methods, including face/eye detection, data rectification, 2D/3D gaze conversion and gaze origin conversion. Finally, we set up a comprehensive benchmark for deep learning-based gaze estimation. We characterize all the public datasets and provide the source code of typical gaze estimation algorithms. This paper serves not only as a reference to develop deep learning-based gaze estimation methods, but also a guideline for future gaze estimation research. The project web page can be found at \emph{\url{https://phi-ai.buaa.edu.cn/Gazehub/}}.

		\end{abstract}
		
		}

	\maketitle

	\IEEEdisplaynontitleabstractindextext

	%
	\IEEEpeerreviewmaketitle
	
	%
	%
	%
	%
	

 \section{Introduction}
	\label{sec_intro}
 \IEEEPARstart{E}{ye} gaze is an essential non-verbal communication cue that contains valuable information about human intent, enabling us to gain insights into human cognition~\cite{Maria_2017_DCN,George_2017_UMAP} and behavior~\cite{Meissner_2019_ORM,Kerr_2019_IJED}.
    Eye gaze has various representations across different applications. Gaze direction serves as the universal representation in most applications. It is defined as a unit direction vector in 3D space originating from eye centers and pointing towards gaze targets. Gaze direction holds significant potential, \eg, in extended reality (XR) devices~\cite{Palazzi_2019_tpami,Sitzmann_2018_tvcg, Wang_2018_Slam}, where it is employed to locate gaze targets in 3D space based on estimated gaze direction.
    By establishing a specific plane in 3D space, gaze direction can be converted into a point of gaze (PoG) on that plane. PoG is widely used in human-computer interaction~\cite{hci_2020_katsini,hci_2018_mohamed,Wang_2015_Hybrid} as it indicates the user's attention area on a screen or display. Additionally, eye gaze can be represented as an attention map in analysis tasks~\cite{lai2022eye,jiang2022does} or as target objects/people in gaze following tasks~\cite{Fang_2021_CVPR,Li_2021_ICCV,Tu_2022_CVPR}.
    Accurate gaze estimation is always crucial for such applications.

    \added{Over the last decades, numerous gaze estimation methods have been proposed.} These methods can be broadly categorized into three groups: 3D eye model recovery-based, 2D eye feature regression-based, and appearance-based methods.
    3D eye model recovery-based methods construct a geometric 3D eye model and estimate gaze directions based on the model.	
    Due to the diversity of human eyes, 3D eye models are usually person-specific.
    The 3D eye model recovery-based methods usually require personal calibration to recover person-specific parameters such as iris radius and kappa angle.
    While these methods often achieve high accuracy, they require dedicated devices such as infrared cameras.
    2D eye feature regression-based methods usually keep the same requirement on devices as 3D eye model recovery-based methods.
    They directly use detected geometric eye feature such as pupil center to regress the point of gaze (PoG).
    They do not require geometric calibration for converting gaze directions into PoG.

    \begin{figure}[t]
		\centering
		\includegraphics[width=\columnwidth]{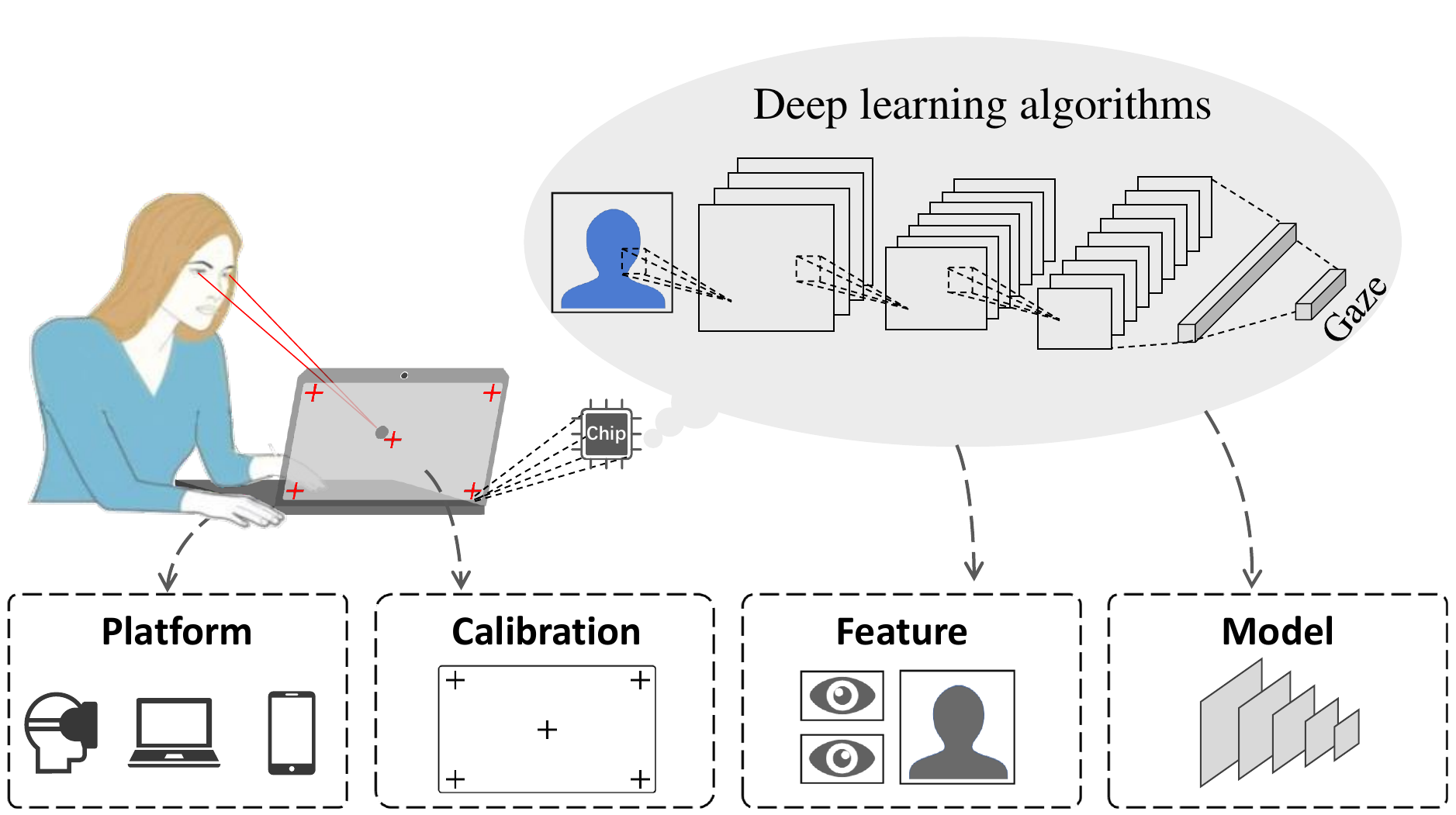} 
		\caption{Deep learning-based gaze estimation relies on simple devices but complex algorithms to estimate human gaze. It usually uses off-the-shelf cameras to capture facial appearance, and employs deep learning algorithms to regress gaze from the appearance. According to this pipeline, we survey current deep learning-based gaze estimation methods from four perspectives: deep feature extraction, deep learning model design, personal calibration, and platforms. }
        \vspace{-5mm}
		\label{Fig:fouraspects}
    \end{figure}

	\begin{figure*}[htp]
		\centering
		\includegraphics[width=1.8\columnwidth]{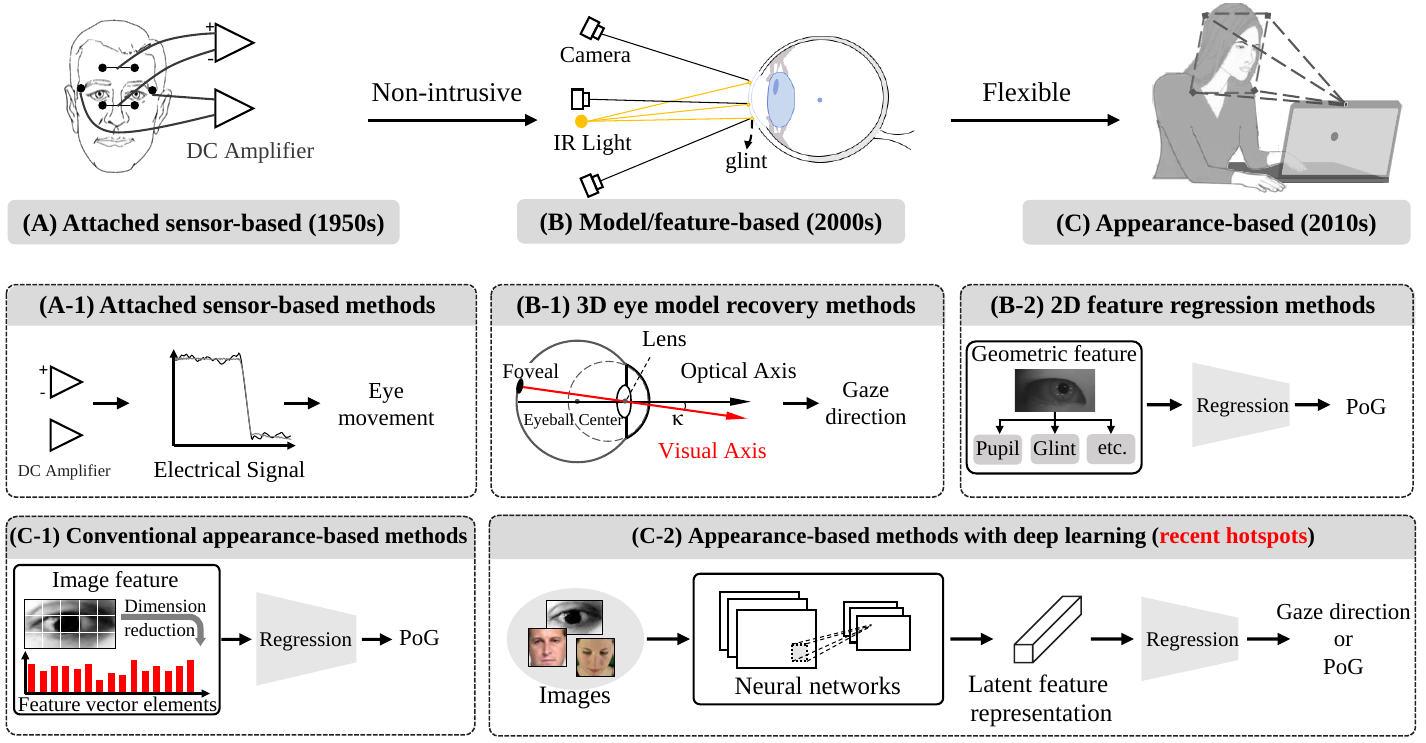} 
		\caption{From intrusive skin electrodes~\cite{Young_1975_survey} to off-shelf web cameras~\cite{Zhang_2015_CVPR}, gaze estimation is more  flexible. Gaze estimation methods are also updated with the change of devices. We illustrate five kinds of gaze estimation methods. (1). Attached sensor-based methods. The method samples the electrical signal of skin electrodes. The signal indicates the eye movement of subjects~\cite{Eggert_2007_No}. (2) 3D eye model recovery methods. The method usually builds a geometric eye model to calculate the visual axis,~\ie, gaze directions. The eye model is fitted based on the light reflection. (3) 2D eye feature regression methods. The method relies on IR cameras to detect geometric eye features such as pupil center, glints, and directly regress the PoG from these features. (4) Conventional appearance-based methods. The method use entire images as feature and directly regress human gaze from features. Some feature reduction methods are also used for extracting low-dimensional feature. For example, Lu~\etal divide eye images into 15 subregion and sum the pixel intensities in each subregion as feature~\cite{Lu_2014_TPAMI}. (5) Appearance-based gaze estimation with deep learning, which is the recent hotspots. Face or eye images are directly inputted into a designed neural network to learn latent feature representation, and human gaze is regressed from the feature representation.}
        \vspace{-5mm}
		\label{Fig:development}
	\end{figure*}

    Appearance-based methods have low device requirements.
    They use off-the-shelf web cameras to capture human eye appearance and regress gaze from the appearance.   
    Although the setup is simple, they have strict requirements on the gaze estimation algorithm.
    They usually require
    1) An effective feature extractor to extract gaze features from high-dimensional raw images.
    Some feature extractors such as histograms of oriented gradients are used in the conventional method~\cite{Martinez_2012_ICIP}. 
    2) A robust regression function to learn the mappings from appearance feature to human gaze. It is non-trivial to map the high-dimensional eye appearance to the low-dimensional gaze. Many regression functions have been used to regress gaze from appearance, \eg, local linear interpolation~\cite{Tan_2002_WACV} and adaptive linear regression~\cite{Lu_2014_TPAMI}. 
    3) A large number of training samples to learn the regression function. They usually collect personal samples with a time-consuming personal calibration, and learn a person-specific gaze estimation model. Some studies seek to reduce the number of training samples 
    ~\cite{Lu_2014_TPAMI}. 
    \vspace{-0.1mm}

    Recently, deep learning-based methods have gained popularity as they offer several advantages over conventional appearance-based methods. These methods use convolution layers or transformers~\cite{vaswani2017attention} to automatically extract high-level gaze features from images. Deep learning models are also highly non-linear and can fit the mapping function from eye appearance to gaze direction even with large head motion. These advantages make deep learning-based methods more accurate and robust than conventional methods.
    Deep learning-based methods also improve cross-subject gaze estimation performance significantly, reducing the need for time-consuming person calibration. These improvements expand the application range of appearance-based gaze estimation.

	In this paper, we provide a systematic review of appearance-based gaze estimation methods using deep learning algorithms.
	As shown in~\fref{Fig:fouraspects}, we discuss these methods from four perspectives: 1) deep feature extraction, 2) deep neural network architecture design, 3) personal calibration, and 4) device and platform.
	From the deep feature extraction perspective, we describe the strategies for extracting features from eye images, face images and videos.
    Under the deep neural network architecture design perspective, we first review methods based on the supervised strategy, containing the supervised, self-supervised, semi-supervised and unsupervised methods.
    Then, We describe different deep neural networks in gaze estimation including multi-task CNNs, recurrent CNNs.
    Furthermore, we introduce methods that integrate CNN models and prior knowledge of gaze.
    From the personal calibration perspective, we describe how to use calibration samples to further improve the performance of CNNs.
    We also introduce the method integrating user-unaware calibration sample collection mechanism.
    Finally, from the device and platforms perspective, we consider different cameras,~\ie, RGB cameras, IR cameras and depth cameras, as well as different platforms,~\ie, computers, mobile devices and head-mount devices.
    We review the advanced methods using these cameras and proposed for these platforms.
    \vspace{-0.2mm}
 
    Besides deep learning-based gaze estimation methods, we also summarize the practices of gaze estimation.
    We first review the data pre-processing methods of gaze estimation including face and eye detection methods and data rectification methods.
    Then, considering various forms of human gaze,~\eg, gaze direction and PoG, we further provide data post-processing methods.
    These methods describe the geometric conversion between various representations of human gaze.
    We also build gaze estimation benchmarks.
    We collect and implement the codes of typical gaze estimation methods, and evaluate them on various datasets.
    For the different kinds of gaze estimation methods, we convert their result for fair comparisons with data post-processing methods.
    Our benchmarks provide comprehensive comparison between state-of-the-art gaze estimation methods.
	
    The paper is organized as follows. \Sref{sec_background} introduces the background of gaze estimation. We introduce the development and category of gaze estimation methods.
    \Sref{sec_method} reviews the state-of-the-art deep learning-based method. 
    In \Sref{sec_dataset}, we introduce the public datasets as well as data pre-processing and post-processing methods.
    We also build the benchmark in this section.	
    In \Sref{sec_conclusion}, we conclude the development of current deep learning-based methods and recommend future research directions.
    This paper can not only serve as a reference to develop deep learning-based gaze estimation methods, but also a guideline for future gaze estimation research.
	
    \section{Gaze Estimation Background}
	\label{sec_background}
	
	\subsection{Categorization}
	\label{ssec_category}
 
        \Fref{Fig:development} illustrates the development of gaze estimation methods. 
        Early gaze estimation methods detect eye movement patterns such as fixation, saccade and smooth pursuit~\cite{Young_1975_survey}. 
	They attach sensors around eyes and measure eye movement using potential differences~\cite{Mowrer_1936_APS, Schott_1922_uber}. 
	With the development of computer vision technology, modern eye-tracking devices have emerged. 
       They usually estimate gaze using eye/face images captured by cameras.
	In general, there are two types of such devices, remote eye tracker and head-mounted eye tracker. The remote eye tracker usually keeps a certain distance from the user, \eg, $\sim$60 cm. The head-mounted eye tracker usually mounts the cameras on a frame of glasses. 
	Compared to the intrusive eye tracking devices, the modern eye tracker greatly enlarges the range of application.
	
	Computer vision-based methods can be further divided into three types: 2D eye feature regression methods, 3D eye model recovery methods and appearance-based methods. 
	The first two types of methods estimate gaze based on geometric features such as contours, reflection and eye corners. The geometric features can be accurately extracted with the assistance of dedicated devices, \eg, infrared cameras. 
	More concretely, the 2D eye feature regression method learns a mapping function from geometric feature to point of gaze, \eg, the polynomials~\cite{Morimoto_2005_CVIU,Stampe_1993_BRMIC} and the neural networks~\cite{Ji_2002_RI}. 	
	The 3D eye model recovery method builds subject-specific geometric eye models to estimate human gaze directions.
	The eye model is fitted with geometric features, such as the infrared corneal reflections~\cite{Guestrin_2006_TBE,Zhu_2007_TBE}, pupil center~\cite{Valenti_2012_TIP} and iris contours~\cite{Mora_2014_CVPR}.	
	However, they usually require a personal calibration process for each subject, since the eye model contains subject-specific parameters such as cornea radius, kappa angles. 
	
	Appearance-based methods directly learn a mapping function from images to human gaze. 
	Different from previous methods, appearance-based methods do not require dedicated devices for detecting geometric features. 
	They use image features such as image pixel~\cite{Lu_2014_TPAMI} or deep features~\cite{Zhang_2015_CVPR} to regress gaze. Various regression models have been used, \eg, neural networks~\cite{Baluja_1994_CMU}, gaussian process regression~\cite{Williams_2006_CVPR}, adaptive linear regression~\cite{Lu_2014_TPAMI}, convolutional neural networks~\cite{Zhang_2015_CVPR} and transformers~\cite{cheng2021gaze}. However, it is still a challenging task due to complex facial appearance.

	\subsection{Appearance-based Gaze Estimation}
	\label{ssec_appbased}
	Appearance-based methods directly learn mapping function from eye appearance to human gaze.	
	As early as 1994, Baluja~\etal propose a neural network and collect 2,000 samples for training~\cite{Baluja_1994_CMU}. 
	Tan~\etal use a linear function to interpolate unknown gaze position using 252 training samples~\cite{Tan_2002_WACV}. 
	These methods usually learn a subject-specific mapping function. 
	They require a time-consuming data collection for the specific subject. To reduce the number of  training samples, Williams~\etal introduce semi-supervised gaussian process regression methods~\cite{Williams_2006_CVPR}. 
	Sugano~\etal propose a method that combines gaze estimation with saliency~\cite{Sugano_2013_TPAMI}. 
	Lu~\etal propose an adaptive linear regression method to select an optimal set of sparsest training sample for interpolation~\cite{Lu_2014_TPAMI}.	
	However, these methods only show reasonable performance in a constrained environment,~\ie, fixed head pose and the specific subject. Their performance significantly degrades when tested on an unconstrained environment. This problem is always challenging in appearance-based gaze estimation.
	
	To address the performance degradation across subjects, Funes~\etal present a cross-subject training method~\cite{Mora_2013_ICIP}. However, the reported mean error is larger than 10 degrees.	Sugano~\etal introduce a learning-by-synthesis method~\cite{Sugano_2014_CVPR}. They use a large number of synthetic cross-subject data to train their model. Lu~\etal employ a sparse auto-encoder to learn a set of bases from eye image patches and reconstruct the eye image using these bases~\cite{Lu_2016_NC}. On the other hand, to tackle the head motion problem, Sugano~\etal cluster the training samples with similar head poses and interpolate the gaze in local manifold~\cite{Sugano_2008_ECCV}. Lu~\etal initiate the estimation with the original training images and compensating for the bias via regression~\cite{Lu_2014_IVC}. They further propose a novel gaze estimation method that handles the free head motion via eye image synthesis using a single camera~\cite{Lu_2015_TIP}.
	
	
	\begin{figure}[t]
		\centering
		\includegraphics[width=0.95\columnwidth]{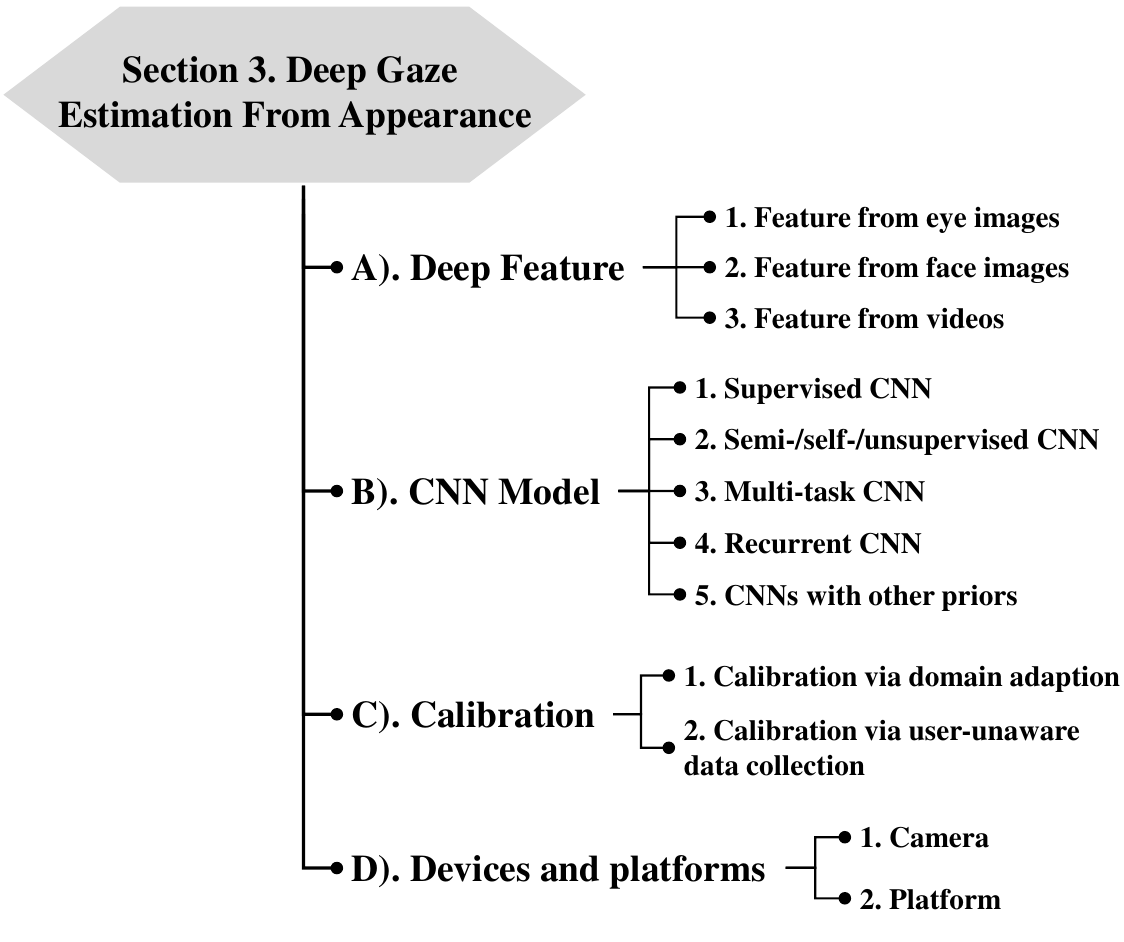} 
		\caption{The architecture of section 3. We introduce gaze estimation with deep learning from four perspectives. }
            \vspace{-5mm}
		\label{fig:structure}
	\end{figure}
	
	\subsection{Deep Learning for Gaze Estimation}
	\label{ssec_dl}
	Appearance-based gaze estimation suffers from many challenges, including head motion and subject differences, particularly in the unconstrained environment. 
	Traditional appearance-based methods often struggle to effectively address these challenges due to their limited fitting ability.
	Deep learning have been used in many computer vision tasks and demonstrated outstanding performance. Zhang~\etal propose the first CNN-based gaze estimation method to regress gaze directions from eye images~\cite{Zhang_2015_CVPR}. They use a simple CNN and the performance surpasses most of the conventional appearance-based approaches.
	Following this study, an increasing number of improvements and extensions on CNN-based gaze estimation methods emerged. Face images~\cite{Krafka_2016_CVPR} and videos~\cite{Kellnhofer_2019_ICCV} have also been used for gaze estimation.
	These inputs provide more valuable information than using eye images alone.
	Some methods are proposed for handling the challenges in an unconstrained environment.
	For example, Cheng~\etal use asymmetric regression to handle the extreme head pose and illumination condition~\cite{Cheng_2020_tip}. 
	Park~\etal learn a pictorial eye representation to alleviate the personal appearance difference~\cite{Park_2018_ECCV}. The calibration-based methods learn a subject-specific CNN model~\cite{Yu_2019_CVPR,Park_2019_ICCV}.
	Xu~\etal investigated the vulnerability of appearance-based gaze estimation~\cite{xu2021vulnerability}.
	
	\section{Deep Gaze Estimation From Appearance}
	\label{sec_method}
	We survey deep learning-based gaze estimation methods in this section.
	We introduce these methods from fours perspectives, deep feature extraction, deep neural network architecture design, personal calibration as well as device and platform.
	\Fref{fig:structure} gives an overview of this section.

	\begin{figure*}[t]
		\centering
		\includegraphics[width=1.95\columnwidth]{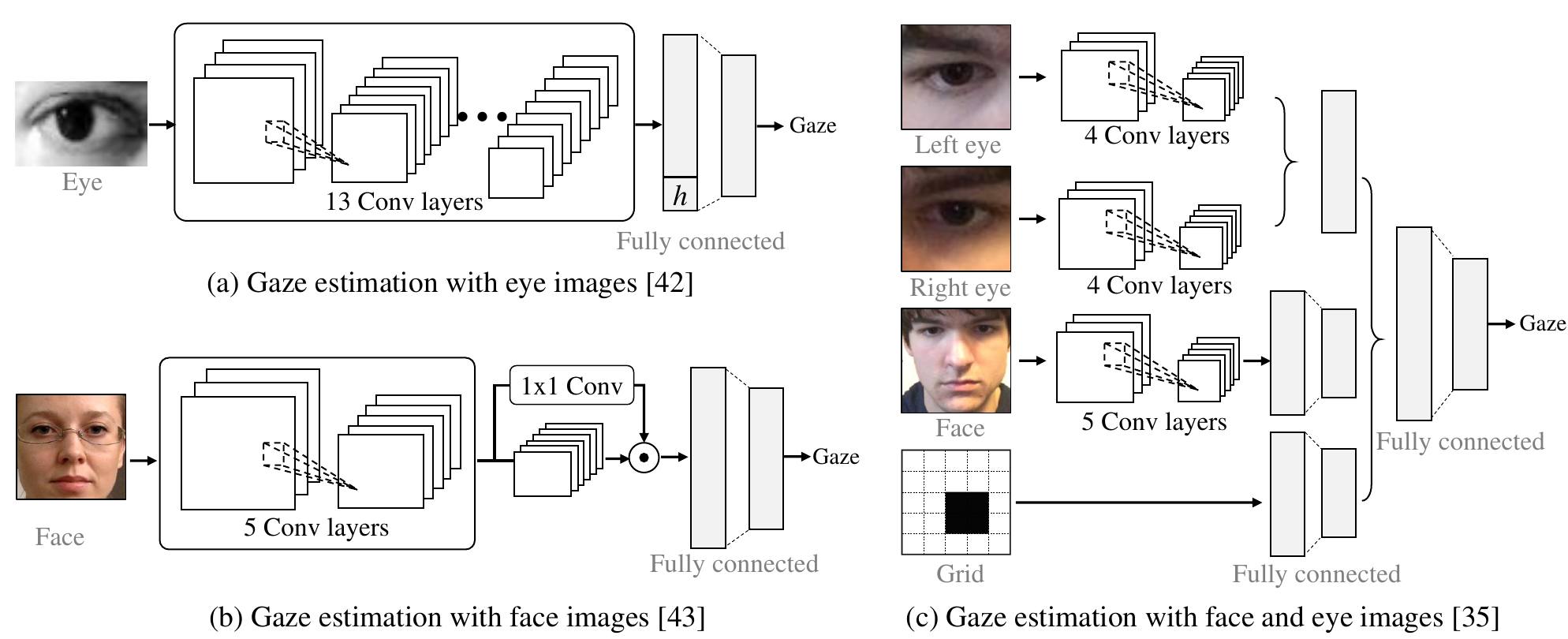} 
		\caption{Some typical CNN-based gaze estimation networks. (a). Gaze estimation with eye images~\cite{Zhang_2017_tpami}. (b) Gaze estimation with face images~\cite{Zhang_2017_CVPRW}. (c). Gaze estimation with face and eye images~\cite{Krafka_2016_CVPR}. }
        \vspace{-5mm}
		\label{fig:feature}
	\end{figure*}
 
	\subsection{Deep Feature from Appearance}
	\label{ssec_feature}
 
	Feature extraction plays a crucial role in most of the learning-based tasks. It is challenging to effectively extract features from complex eye appearance due to identity, illumination and \textit{etc.} The quality of the extracted features determines the gaze estimation accuracy. In this section, we summarize feature extraction mechanisms according to the types of input into the deep neural network, including eye images, face images and videos.
 
	
	\subsubsection{Feature from Eye Images}
	\label{sssec_featureeye}
    
    Human gaze has a strong correlation with eye appearance. Even a minor perturbation in gaze direction can result in noticeable changes in eye appearance. For instance, when the eyeball rotates, the position of the iris and the shape of the eyelid undergo alterations, leading to corresponding changes in gaze direction. This relationship between gaze and eye appearance enables the gaze estimation based on the visual feature of eyes.
    Conventional methods typically estimate gaze using high-dimensional raw image features~\cite{Tan_2002_WACV,Xu_1998_BMVC}. These features are obtained by raster scanning all the pixels in eye images, resulting in a representation that contains a significant amount of redundancy. Moreover, these features are highly sensitive to environmental changes, which can pose challenges in achieving accurate gaze estimation.

	Deep learning-based methods automatically extract deep features from eye images. 
	Zhang \etal propose the first deep learning-based gaze estimation method~\cite{Zhang_2015_CVPR}. 
	They employ a CNN to extract features from grey-scale eye images and concatenate the features with head pose. 
	As with most deep learning tasks, the deeper network structure and larger receptive field, the more informative features can be extracted. 
	Zhang~\etal~\cite{Zhang_2017_tpami} further extend their previous work~\cite{Zhang_2015_CVPR} and present a GazeNet which is inherited from a 16-layer VGG network \cite{Simonyan_2014_arxiv}.
	Chen~\etal~\cite{Chen_2019_ACCV} use dilated convolutions to extract high-level eye features, which efficiently increases the receptive field size of the convolutional filters without reducing spatial resolution.
	 
	Recent studies found that concatenating the features of two eyes helps to improve the gaze estimation accuracy \cite{Fischer_2018_ECCV,Cheng_2018_ECCV}. 
	Fischer~\etal~\cite{Fischer_2018_ECCV} employ two VGG-16 networks to extract individual features from two eye images, and concatenate two eye features for regression. 
	Cheng \etal~\cite{Cheng_2018_ECCV} build a four-stream CNN network for extracting features from two eye images. 
	Two streams of CNN are used for extracting individual features from left/right eye images, the other two streams are used for extracting joint features of two eye images. 
	They claim that the two eyes are asymmetric, and  propose an asymmetric regression and evaluation network to extract different features from two eyes. 
	More recent studies propose to use attention mechanism to fuse two eye features. Cheng~\etal~\cite{Cheng_2020_AAAI} argue that the weights of two eye features are determined by face images due to the specific task in \cite{Cheng_2020_AAAI}, so they assign weights for two eye features with the guidance of facial features. 
	Bao~\etal~\cite{Bao_2020_ICPR} propose a self-attention mechanism to fuse two eye features. They concatenate the feature maps of two eyes and use a convolution layer to generate the weights of the feature map.
	Murthy~\etal~\cite{biswas2021appearance} simultaneously estimate feature vectors and weights for each eye image and concatenate the left and right eye feature. They also propose a network which obtains the difference between left and right eye feature to circumvent person-dependent features.
	
	Above methods extract the general features from eye images while other works explored extracting specific features to handle the head motion and subject difference. Several studies have attempted to extract subject-invariant features from eye images~\cite{Park_2018_ECCV,Wang2_2019_CVPR,Park_2019_ICCV}. 
	Park \etal~\cite{Park_2018_ECCV} convert the original eye images into a unified gaze representation, which is a pictorial representation of eyeball, iris and pupils. 
	They regress gaze directions from the pictorial representation. 
	Wang \etal propose an adversarial learning approach to extract the domain/person-invariant feature \cite{Wang2_2019_CVPR}. They feed the features into an additional classifier and design an adversarial loss function to handle the appearance variations. 
	Park \etal use an autoencoder to learn the compact latent representation of gaze, head pose and appearance~\cite{Park_2019_ICCV}. 
	They introduce a geometric constraint on gaze representations, \ie, the rotation matrix between the two given images transforms the gaze representation of one image to another.
	In addition, some methods use generative adversarial networks (GAN) to pre-process eye images to handle specific environment factors.
	Kim~\etal~\cite{kim_2020_ETRA} convert low-light eye images into bright eye images.
	Rangesh~\etal~\cite{rangesh_2020_IV} use a GAN to remove eyeglasses.
	
	Besides the supervised approaches for extracting gaze features, unannotated eye images have also been used for learning gaze representations. 
	Yu \etal input the difference of gaze representations from two eyes into pre-trained network for gaze redirection~\cite{Yu_2020_CVPR}. 
	They learn 2-D representations from unannotated eye images which can be seemed as unaligned gaze.
	Sun \etal propose a cross encoder to disentangle gaze feature and appearance feature. They improve the few-shot performance using learned gaze feature.

	\subsubsection{Feature from Face Images}
	\label{sssec_featurefacial}
	Face images contain head pose information that also contributes to gaze estimation.	
	Conventional methods extract features such as head pose~\cite{Lu_2015_TIP} and facial landmarks~\cite{Yamazoe_2008_ETRA, chen_2008_ICPR, Jeni_2016_CVPRW} from face images. 
	Eye image-based methods typically use head pose vectors as an additional input~\cite{Zhang_2015_CVPR,Cheng_2018_ECCV}. 
	\added{Nevertheless, the impact of head pose appears to be marginal~\cite{Zhang_2017_tpami}, particularly when the basic network has already achieved high accuracy. One possible rationale for this observation lies in the fact that a single head pose often corresponds to a broad gaze range~\cite{Zhang_2020_ECCV,park_2020_eccv}, thereby providing only a coarse indication of gaze direction rather than precise gaze information. } 
    Deep facial feature performs better than the head pose.
	Recent studies directly use face images as input and employ a CNN to extract deep facial features~\cite{Zhang_2017_CVPRW, Krafka_2016_CVPR, mishra2020360,zhang2022gazeonce} as shown in~\fref{fig:feature} (b).
	It demonstrates an improved performance than the approaches that only use eye images.
    Cheng \etal~\cite{cheng2021gaze} explore the transformer for gaze estimation. They use CNN to extract feature maps from face images and input the feature map into transformer encoder for gaze estimation.

	Face images contain redundant information. 
	Researchers have attempted to filter out the useless features in face image~\cite{Zhang_2017_CVPRW, Ogusu_2019_FG}. 
	Zhang \etal~\cite{Zhang_2017_CVPRW} propose a spatial weighting mechanism to efficiently encode the location of the face into a standard CNN architecture.
	The system learns spatial weights based on the activation maps of the convolutional layers. 
	This helps to suppress the noise and enhance the contribution of the highly activated regions. Zhang \etal~\cite{Zhang_2020_BMVC} propose a learning-based region selection method by dynamically selecting suitable sub-regions from facial region. Cheng \etal~\cite{cheng_2022_aaai} propose a plug-and-play self-adversarial network to purify facial features. They remove gaze-irrelevant image features while preserve gaze-relevant features, so the robustness of gaze estimation network has been improved.

	Some studies crop the eye image out of the face images and directly feed it into the network. 
	These works usually use a three-stream network to extract features from face images, left and right eye images, respectively as shown in ~\fref{fig:feature} (c)~\cite{Krafka_2016_CVPR,Chen_2019_ACCV,Yu_2020_ICMI,zhang_2020_ICBSIC,wang_2020_WACV}. 
	Besides, Deng \etal~\cite{Deng_2017_ICCV} decompose gaze directions into the head rotation and eyeball rotation. 
	They use face images to estimate the head rotation and eye images to estimate the eyeball rotation. 
	These two rotations are aggregated into a gaze vector through a gaze transformation layer. 
	Cheng \etal~\cite{Cheng_2020_AAAI} propose a coarse-to-fine gaze estimation method.  
	They use face feature to estimate basic gaze directions, then refine the basic gaze direction with eye features. 
	They use GRU~\cite{Cho_2014_arxiv} to build the network.
	Cai \etal~\cite{cai2021gaze} use a transformer encoder~\cite{vaswani2017attention} to aggregate face and eye features. They feed face and two eye features into the transformer encoder and concatenate the outputs of the encoder for gaze estimation.

	\begin{figure}[t]
		\centering
		\includegraphics[width=1.0\columnwidth]{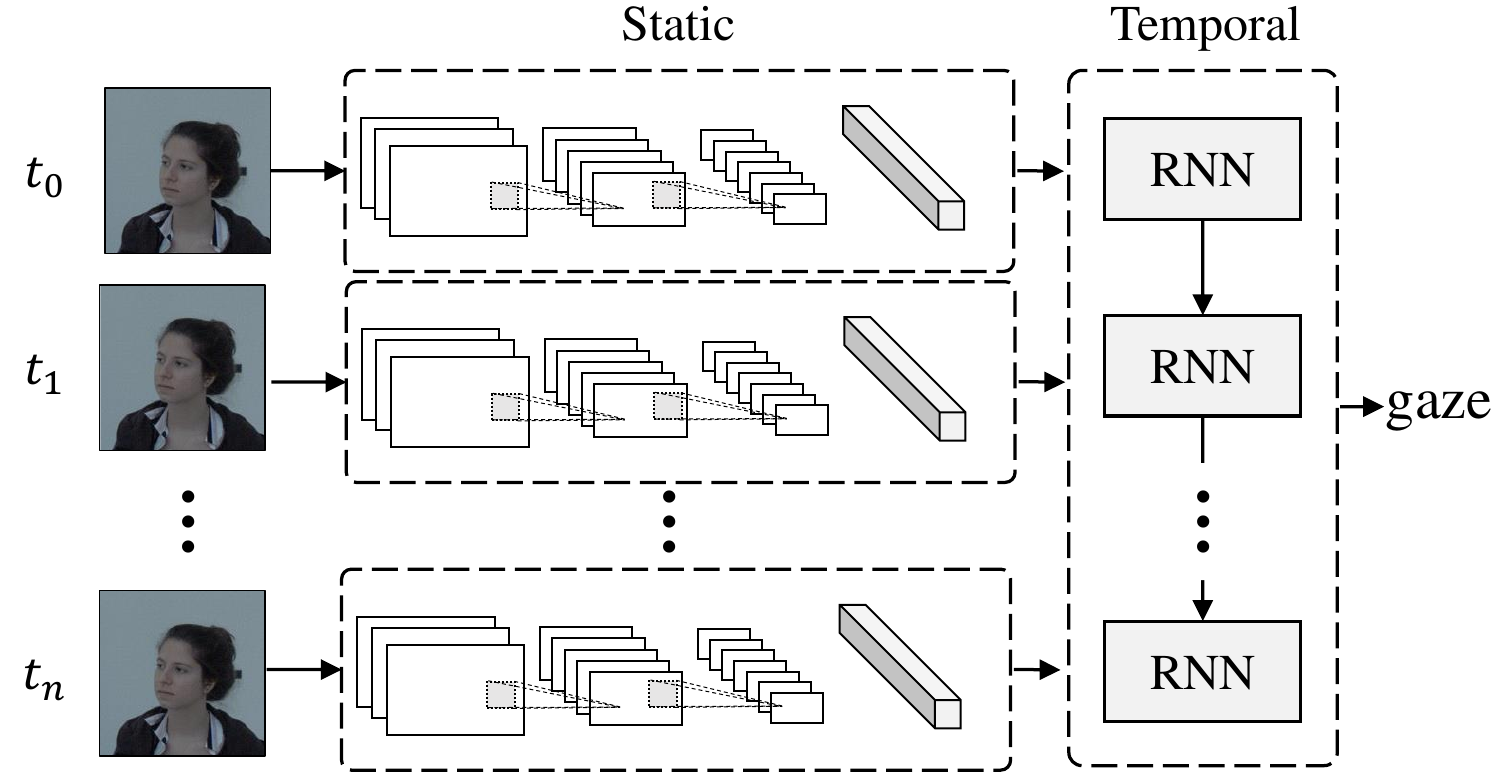} 
		\caption{Gaze estimation with videos. It first extracts static features from each frame using a typical CNN, and feeds these static features into RNN for extracting temporal information. }
            \vspace{-5mm}
		\label{fig:rcnn}
	\end{figure}
	
	Facial landmarks have also been used as additional features to model the head pose and eye position. 
	Palmero \etal combine individual streams (face, eyes region and face landmarks) in a CNN~\cite{Palmero_2018_BMVC}. 
	Dias \etal extract the facial landmarks and directly regress gaze from the landmarks~\cite{Dias_2020_WACV,her2023uncertainty}. The network outputs the gaze direction as well as an estimation of its own prediction uncertainty. Jyoti \etal further extract geometric features from the facial landmark locations~\cite{Jyoti_2018_icpr}. 
	The geometric feature includes the angles between the pupil center as the reference point and the facial landmarks of the eyes and the tip of the nose.  
    The detected facial landmarks can also be used for unsupervised gaze representation learning.	Dubey \etal~\cite{Dubey_2019_IJCNN} collect the face images from the web and annotate their gaze zone based on the detected landmarks. They perform gaze zone classification tasks on the dataset for unsupervised gaze representation learning. In addition, since the cropped face image does not contain face position information, Krafka \etal~\cite{Krafka_2016_CVPR} propose the iTracker, combining the information from left/right eye images, face images as well as face grid information. The face grid indicates the position of the face region in images and it is usually used in PoG estimation.
	
	\subsubsection{Feature from Videos}
	\label{sssec_featurevideo}
	Temporal information from videos also contributes to better gaze estimates. Recurrent Neural Network (RNN) has been widely used in video processing, \eg, long short-term memory (LSTM)~\cite{Kellnhofer_2019_ICCV,Zhou_2019_ICME}. 
	As shown in~\fref{fig:rcnn}, they usually use a CNN to extract features from face images at each frame, and then input these features into a RNN.
    The temporal relations between each frame are automatically captured by the RNN for gaze estimation.
 
	Temporal features such as the optical flow and eye movement dynamics have been used to improve gaze estimation accuracy. 
	The optical flow provides the motion information between the frames. 
	Wang \etal~\cite{Wang_2019_tvcg} use the optical flow constraints with 2D facial features to reconstruct the 3D face structure based on the input video frames. 
	Eye movement dynamics, such as fixation, saccade and smooth pursuits, have also been used to improve gaze estimation accuracy. 
	Wang \etal~\cite{Wang_2019_CVPR} propose to leverage eye movement to generalize eye tracking algorithm to new subjects. 
	They use a dynamic gaze transition network to capture underlying eye movement dynamics and serve as prior knowledge. 
	They also propose another static gaze estimation network, which estimates gaze based on the static frame. 
	They finally combine the two networks for better gaze estimation accuracy.
        The combination method of the two networks is solved as a standard inference problem of linear dynamic system or Kalman filter~\cite{Murphy_2002_book}.
	
	
	\subsection{CNN Models}
	\label{ssec_models}
	Convolutional neural networks have been widely used in many compute vision tasks~\cite{Farabet_2012_PAMI}. They also demonstrate superior performance in the field of gaze estimation. 
	In this section, we first review the existing gaze estimation methods from the learning strategy perspective, \ie, the supervised CNNs and the semi-/self-/un-supervised CNNs. Then we introduce the different network architectures,\ie, multi-task CNNs and the recurrent CNNs for gaze estimation.	
	In the last part of this section, we discuss the CNNs that integrate prior knowledge to improve performance.
	
	\subsubsection{Supervised CNNs}
	\label{sssec_supcnn}
	
	Supervised CNNs are the most commonly used networks in appearance-based gaze estimation~\cite{Zhang_2015_CVPR, liu_2020_iccpr,mahanama_2020_AH,lemley_2019_TCE}. 
	\fref{fig:feature} shows the typical architecture of supervised gaze estimation CNN. 
	The network is trained using image samples with ground truth gaze directions. 
	The gaze estimation problem is essentially learning a mapping function from raw images to human gaze. 
	Therefore, similar to other computer vision tasks~\cite{He_2016_CVPR}, the deeper CNN architecture usually achieves better performance.
	A number of CNN architectures that have been proposed for typical computer vision tasks also show great success in gaze estimation task,~\eg,	LeNet~\cite{Zhang_2015_CVPR}, AlexNet~\cite{Zhang_2017_CVPRW}, VGG~\cite{Zhang_2017_tpami}, ResNet18~\cite{Kellnhofer_2019_ICCV} and ResNet50~\cite{Zhang_2020_ECCV}. 
	Besides, some well-designed modules also help to improve the estimation accuracy~\cite{Chen_2019_ACCV,Cheng_2020_AAAI,zhuang_2021_IAEAC,bublea_2020_ISETC}. Chen~\etal use a dilated convolution to extract features from eye images~\cite{Chen_2019_ACCV}. Cheng~\etal propose an attention module for fusing two eye features~\cite{Cheng_2020_AAAI}. Cheng~\etal integrate the CNN and the transformer encoder~\cite{vaswani2017attention} to improve the estimation performance~\cite{cheng2021gaze}.
	
	
	To supervise the CNN during training, the system requires the large-scale labeled dataset. 
	Several large-scale datasets have been proposed~\cite{Zhang_2015_CVPR,Krafka_2016_CVPR}. 
	However, it is difficult and time-consuming to collect enough gaze data in practical applications. 
	Inspired by the physiological eye model~\cite{Ruhland_2014_Eurographics}, some researchers propose to synthesize labeled photo-realistic image~\cite{Sugano_2014_CVPR, Swirski_2014_etra, Porta_2019_iccvw}. 
	These methods usually build eye-region models and render new images from these models. 
	One of such methods is proposed by Sugano~\etal~\cite{Sugano_2014_CVPR}. 
	They synthesize dense multi-view eye images by recovering the 3D shape of eye regions, where they use a patch-based multi-view stereo algorithm~\cite{Furukawa_2009_PAMI} to reconstruct the 3D shape from eight multi-view images. 
	Wood~\etal propose to synthesize the close-up eye images for a wide range of head poses, gaze directions and illuminations to develop a robust gaze estimation algorithm~\cite{Wood_2015_ICCV}.
	Following this work, Wood~\etal further propose another system named UnityEye to rapidly synthesize large amounts of eye images of various eye regions ~\cite{Wood_2016_etra}. 
	To make the synthesized images more realistic, Shrivastava \etal propose an unsupervised learning paradigm using generative adversarial networks to improve the realism of the synthetic images~\cite{Shrivastava_2017_CVPR}. 
    Wang~\etal plot eye shapes based on geometric models and use GAN to render eye images~\cite{Wang_2018_CVPR}.
    These methods serve as data augmentation tools to improve the performance of gaze estimation.
	
	Gaze redirection has also been used as a data augmentation tool. It generates face images with target gaze based on given face images. Recently, many gaze redirection methods have been proposed~\cite{jindal2021cuda, yu2019improving, He_2019_ICCV, zheng2020self} and bring significant performance improvement. NeRF~\cite{mildenhall2020nerf} shows great multi-view consistency and is used to learn implicit face model from multi-view images. It can also renders face images under novel gaze for gaze redirection~\cite{ruzzi2023gazenerf,yin2022nerf}.
	
	
	
	
	\begin{figure}[t]
		\centering
		\includegraphics[width=0.95\columnwidth]{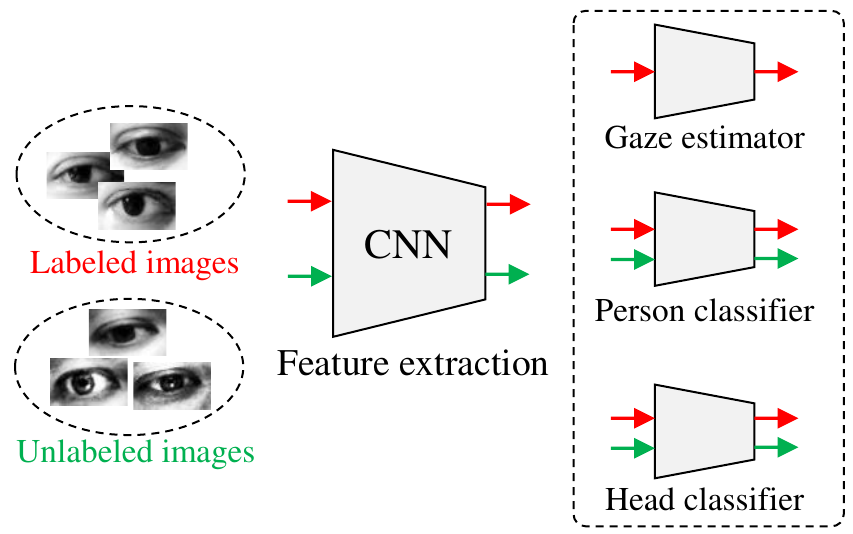} 
		\caption{A semi-supervised CNN~\cite{Wang2_2019_CVPR}. It uses both labeled images and unlabeled images for training. It designs an extra appearance classifier and a head pose classifier. The two classifiers align the feature of labeled images and unlabeled images.}
		\label{fig:semisup}
	\end{figure}
	
	\begin{figure}[t]
		\centering
		\includegraphics[width=0.95\columnwidth]{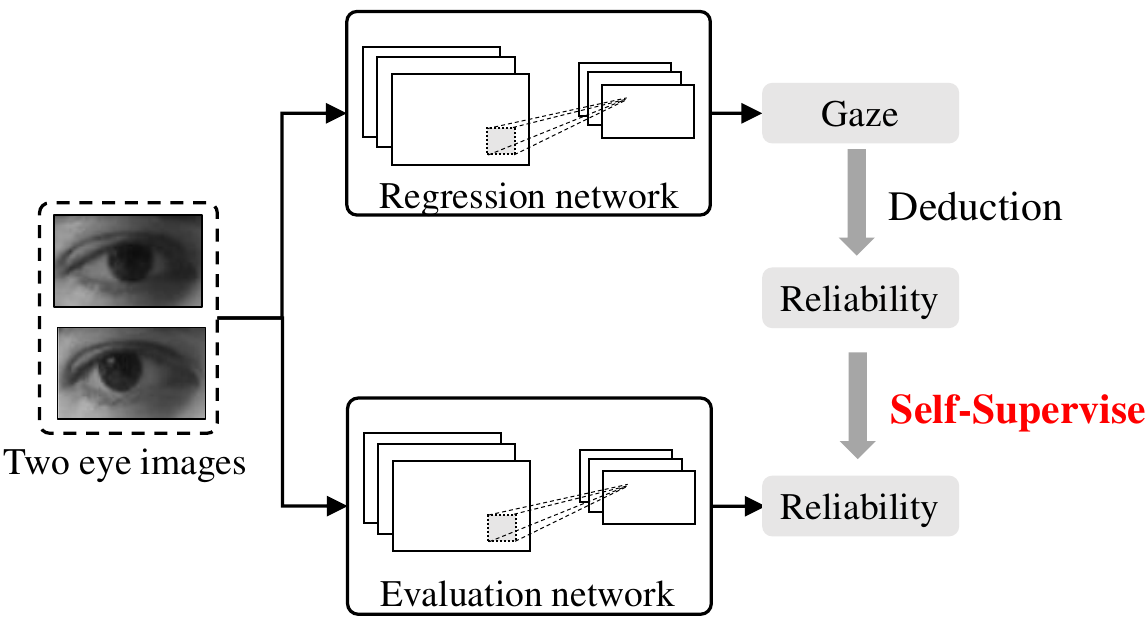} 
		\caption{A self-supervised CNN~\cite{Cheng_2018_ECCV}. The network is consisted of two sub-networks. The  regression network estimates gaze from two eye images and generates the ground truth of the other network for self-supervision. }
            \vspace{-5mm}
		\label{fig:selfsup}
	\end{figure}

	\begin{figure}[t]
		\centering	
		\includegraphics[width=0.95\columnwidth]{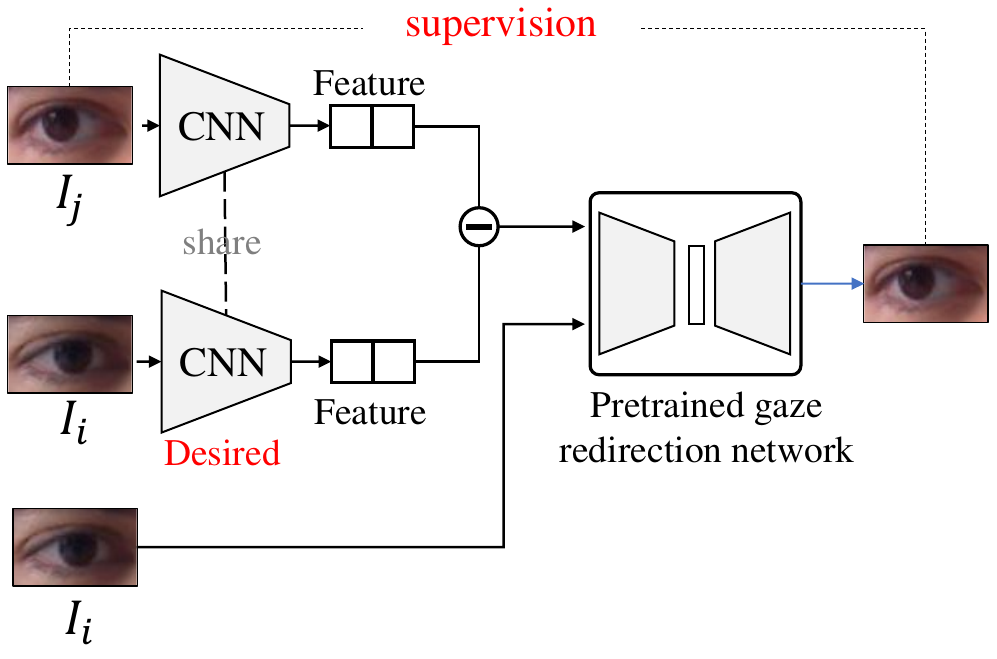} 
		\caption{An unsupervised CNN~\cite{Yu_2020_CVPR}. \added{It extracts 2D feature from eye images. The feature difference and one eye image are fed into a pretrained gaze redirection network to generate the other eye image.} }
           \vspace{-5mm}
		\label{fig:unsup}
	\end{figure}

	\subsubsection{Semi-/Self-/Un-supervised CNNs}
	\label{sssec_unsupcnn}
	
    
	Semi-/self-/un-supervised CNNs attract much attention recently and also show large potential in gaze estimation.
    There are typically two main topics in recent research. 1) Gaze data collection is time-consuming and expensive. To reduce the requirement on annotated images, some methods leverage unannotated images to learn robust feature representation~\cite{kothari2021weakly,Yu_2020_CVPR}.
    2) Gaze estimation methods show performance drop in new environments/domains. Researchers use annotated images in source domains and unannotated images in target domains to improve the performance in target domains~\cite{Bao_2022_CVPR,liu2021generalizing}.
    The second topic is more systematic than the first topic with recent development. It is defined as unsupervised domain adaption, where the ``unsupervised" aspect refers to the lack of labelled data in the target domain.

	Semi-supervised CNNs require both labeled and unlabeled images for optimizing networks.	
	Wang \etal propose an adversarial learning approach to improve the model performance on the target subject/dataset~\cite{Wang2_2019_CVPR}. 
	As shown in~\fref{fig:semisup}, it requires labeled images in the training set as well as unlabeled images of the target subject/dataset. 
    They use the labeled data to supervise the gaze estimation network and design an adversarial module for semi-supervised learning.
	Given these features used for gaze estimation, the adversarial module tries to distinguish their source and the gaze estimation network aims to extract subject/dataset-invariant features to cheat the module.
	Kothari \etal~\cite{kothari2021weakly} found the strong gaze-related geometric constraints when people "look at each other" (LAEO). They estimate 3D and 2D landmarks in the images of LAEO dataset~\cite{Marin-Jimenez_2019_CVPR}, and generate pseudo gaze annotation for gaze estimation. While it cannot bring competitive performance, therefore, they further integrate labeled images and LAEO datasets for semi-supervised gaze estimation.
	
	Self-supervised CNNs aim to formulate a pretext auxiliary learning task to improve the estimation performance. 
	Cheng~\etal~propose a self-supervised asymmetry regression network for gaze estimation~\cite{Cheng_2018_ECCV}. 
	As shown in~\fref{fig:selfsup}, the network consists of a regression network to estimate the two eyes' gaze directions and an evaluation network to assess the reliability of two eyes.
	During training, the result of the regression network is used to supervise the evaluation network, the accuracy of the evaluation network determines the learning rate in the regression network. 
	They simultaneously train the two networks and improve the regression performance without additional inference parameters. 
	Xiong~\etal introduce a random effect parameter to learn the person-specific information in gaze estimation~\cite{Xiong_2019_CVPR}. 
	They utilize the variational expectation-maximization algorithm~\cite{Beal_2003_variational} and stochastic gradient descent~\cite{robbins_1951_math} to estimate the parameters of the random effect network during training. They use another network to predict the random effect based on the feature representation of eye images. The self-supervised strategy predicts the random effects to enhance the accuracy for unseen subjects. He~\etal introduce a person-specific user embedding mechanism\cite{He_2019_iccvw}. They concatenate the user embedding with appearance features to estimate gaze. They also build a teacher-student network, where the teacher network optimizes the user embedding during training and the student network learns the user embedding from the teacher network.
	
	Unsupervised CNNs only require unlabeled data for training. Nevertheless, it is hard to optimize CNNs without the ground truth. 
	Many specific tasks are designed for unsupervised CNNs. 
	Dubey \etal~\cite{Dubey_2019_IJCNN} collect unlabeled facial images from webpages. 
	They roughly annotate the gaze region based on the detected landmarks. Therefore, they can perform the classical supervised task for gaze representation learning. 
	Yu \etal utilize a pre-trained gaze redirection network to perform unsupervised gaze representation learning~\cite{Yu_2020_CVPR}.  
	As shown in~\fref{fig:unsup}, they use the gaze representation difference of the input and target images as the redirection variables. 
	Given the input image and the gaze representation difference, the gaze network aims to reconstructs the target image. 
	Therefore, the reconstruction task supervises the optimization of the gaze representation network. 
    Sun \etal propose a cross-encoder for unsupervised learning~\cite{Sun_2021_ICCV}. They acquire paired eye images for training where the paired images have the same gaze or appearance. They use an encoder extract appearance and gaze feature from eye images. They exchange the two features of selected paired images and aim to reconstruct the original image based on the exchanged feature.
	Note that, these approaches learn the gaze representation, but they also require a few labeled samples to fine-tune the final gaze estimator.

	
	\begin{figure}[t]
		\centering
		\includegraphics[width=1.0\columnwidth]{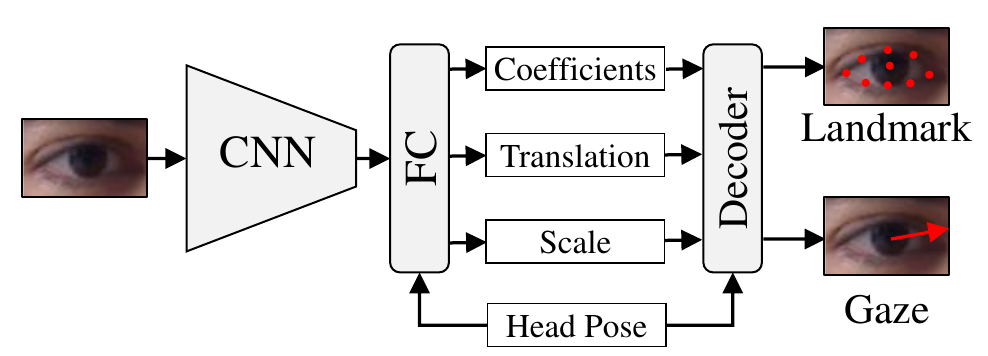} 
		\caption{A multitask CNN~\cite{Yu_2018_ECCVW}. \added{It estimates the coefficients of a landmark-gaze model as well as the scale and translation parameters.} The three results are used to calculate eye landmarks and estimated gaze.  }
        \vspace{-5mm}
		\label{fig:multitask}
	\end{figure}
	
	\subsubsection{Multi-task CNNs}
	\label{sssec_mtcnn}
	Multi-task learning usually contains multiple tasks that provide related domain information as inductive bias to improve model generalization~\cite{Ruder_2017_arxiv}. 
	Some auxiliary tasks are proposed for improving model generalization in gaze estimation. 
	
	Lian \etal propose a multi-task multi-view network for gaze estimation~\cite{Lian_2019_TNNLS}. 
	They estimate gaze directions based on single-view eye images and PoG from multi-view eye images. 
	They also propose another multi-task CNN to estimate PoG using depth images\cite{Lian_2019_AAAI}. 
	They design an additional task to leverage facial features to refine depth images. 
	The network produces four features for gaze estimation, which are extracted from the facial images, the left/right eye images and the depth images.
	
	Some works seek to decompose the gaze into multiple related features and construct multi-task CNNs to estimate these feature.	
	Yu \etal introduce a constrained landmark-gaze model for modeling the joint variation of eye landmark locations and gaze directions~\cite{Yu_2018_ECCVW}. As shown in~\fref{fig:multitask}, they build a multi-task CNN to estimate the coefficients of the landmark-gaze model as well as the scale and translation information to align eye landmarks. Finally, the landmark-gaze model serve as a decode to calculate gaze from estimated parameters.. Deng~\etal decompose the gaze direction into eyeball movement and head pose~\cite{Deng_2017_ICCV}. They design a multi-tasks CNN to estimate the eyeball movement from eye images and the head pose from facial images. 
	The gaze direction is computed from eyeball movement and head pose using geometric transformation. Wu \etal propose a multi-task CNN that simultaneously segments the eye part, detects the IR LED glints, and estimates the pupil and cornea center \cite{Wu_2019_iccvw}. 
	The gaze direction is covered from the reconstructed eye model.
	
	Other works perform multiple gaze-related tasks simultaneously. 
	Recasens \etal present an approach for following gaze in video by predicting where a person (in the video) is looking, even when the object is in a different frame \cite{Recasens_2017_ICCV}. 
	They build a CNN to predict the gaze location in each frame and the probability containing the gazed object of each frame. Also, visual saliency shows strong correlation with human gaze in scene images~\cite{Kruthiventi_2016_CVPR, Wang_2020_tpami}. In~\cite{Chong_2018_ECCV}, they estimate the general visual attention and human's gaze directions in images at the same time. Kellnhofer \etal propose a temporal 3D gaze network~\cite{Kellnhofer_2019_ICCV}. They use bi-LSTM~\cite{Graves_2005_ICANN} to process a sequence of 7 frames to estimate not only  gaze directionS but also gaze uncertainty.
	
	
	
	\subsubsection{Recurrent CNNs}
	\label{sssec_recurcnn}
	
	Human eye gaze is continuous. This inspires researchers to improve gaze estimation performance by using temporal information. Recently, recurrent neural networks have shown great capability in handling sequential data. Some researchers employ recurrent CNNs to estimate the gaze in videos~\cite{Palmero_2018_BMVC,Kellnhofer_2019_ICCV,Zhou_2019_ICME}. 
	
	We first give a typical example of the data processing workflow. 
	Given a sequence of frames $\{X_1, X_2, ..., X_N\}$, a united CNN $f_U$ is used to extract feature vectors from each frame,~\ie,~$x_t = f_U(X_t)$. These feature vectors are fed into a recurrent neural network $f_R$ and the network outputs the gaze vector,~\ie,~$g_i = f_R(x_1, x_2, ..., x_N)$.
    Palmero~\etal set $N=4$ and $i=4$ in their method. They input four frames to estimate the gaze of the last frame~\cite{Palmero_2018_BMVC}.
    Kellnhofer~\etal set $N=7$ and $i=4$~\cite{Kellnhofer_2019_ICCV}. 
    They consider extra three frames after the target frame compared with Palmero.
    These methods both select the nearest three frames (including the previous and the next three frames) for additional vision feature. Besides, some methods utilize the past gaze trajectory for gaze prediction\cite{Xu_2018_CVPR, Wang_2019_CVPR}. They select a larger time range, \eg, $1\sim2s$ ($30\sim90$ frames), in the gaze prediction task. 
    We visualize a expample network architecture in~\fref{fig:rcnn}.

	Different types of input have been explored to extract features. Kellnhofer \etal directly extract features from facial images~\cite{Kellnhofer_2019_ICCV}. Zhou \etal combine the feature extracted from facial and eye images~\cite{Zhou_2019_ICME}. Palmero \etal use facial images, binocular images and facial landmarks to generate the feature vectors~\cite{Palmero_2018_BMVC}. 
	Different RNN structures have also been explored, such as GRU~\cite{Cho_2014_arxiv} in ~\cite{Palmero_2018_BMVC}, LSTM~\cite{Hochreiter_1997_NC} in~\cite{Zhou_2019_ICME} and bi-LSTM~\cite{Graves_2005_ICANN} in~\cite{Kellnhofer_2019_ICCV}. Cheng~\etal~leverage the recurrent CNN to improve gaze estimation performance from static images rather than videos~\cite{Cheng_2020_AAAI}. They generalize the gaze estimation as a sequential coarse-to-fine process and use GRU to relate the basic gaze direction estimated from facial images and the gaze residual estimated from eye images.
	

	\subsubsection{CNNs with Other Priors}
	\label{sssec_othercnn}
	Prior information also helps to improve gaze estimation accuracy, such as decomposition of gaze direction, anatomical eye models and eye movement patterns~\cite{Deng_2017_ICCV, Cheng_2018_ECCV, Park_2018_ECCV, Wang_2019_CVPR, Xiong_2019_CVPR, Chen_2020_WACV}.
	
	\emph{Decomposition of Gaze Direction.} Human gaze can be decomposed into the head pose and the eyeball pose. Deng~\etal use two CNNs to estimate head pose from facial images and eyeball pose from eye images. They integrate these two results into final gaze directions using geometric transformation \cite{Deng_2017_ICCV}.
	
	\emph{Anatomical Eye Model.} The human eye is composed of eyeball, iris, pupil center and \etc~Park~\etal propose a pictorial gaze representation based on the eye model~\cite{Park_2018_ECCV}. They render the eye model to generate a pictorial image, where the pictorial image eliminates the appearance variance. They first generate pictorial images from original images using CNN and use another CNN to estimate gaze directions from pictorial image.
	
	\emph{Eye Movement Pattern.} Common eye movements, such as fixation, saccade and smooth pursuits, are independent of viewing contents and subjects. Wang \etal propose to incorporate the generic eye movement pattern in dynamic gaze estimation \cite{Wang_2019_CVPR}. They recover the eye movement pattern from videos and use a CNN to estimate gaze from static images.
	
	\emph{Two eye asymmetry Property.}
	Cheng \etal discover the 'two eye asymmetry' property that the appearances of two eyes are different while the gaze directions of two eyes are approximately the same~\cite{Cheng_2020_tip}. Based on this observation, Cheng \etal propose to treat two eyes asymmetrically in the CNN. They design an asymmetry regression network for adaptively weighting two eyes. 
	
	\emph{Gaze data distribution.}
	The basic assumption of regression models is independent identically distributed, however, gaze data is not \emph{i.i.d}. Xiong \etal discuss the problem~\cite{Xiong_2019_CVPR} and design a mixed-effect model to consider person-specific information.
	
	\emph{Inter-subject bias.}
	Chen \etal observe the inter-subject bias in most datasets \cite{Chen_2020_WACV,chen2022towards}. They make the assumption that there exists a subject-dependent bias that cannot be estimated from images. Thus, they propose a gaze decomposition method. They decompose the gaze into the subject-dependent bias and the subject-independent gaze estimated from images. During test, they use some image samples to calibrate the subject-dependent bias.

	\begin{figure}[t]
		\centering
		\includegraphics[width=0.95\columnwidth]{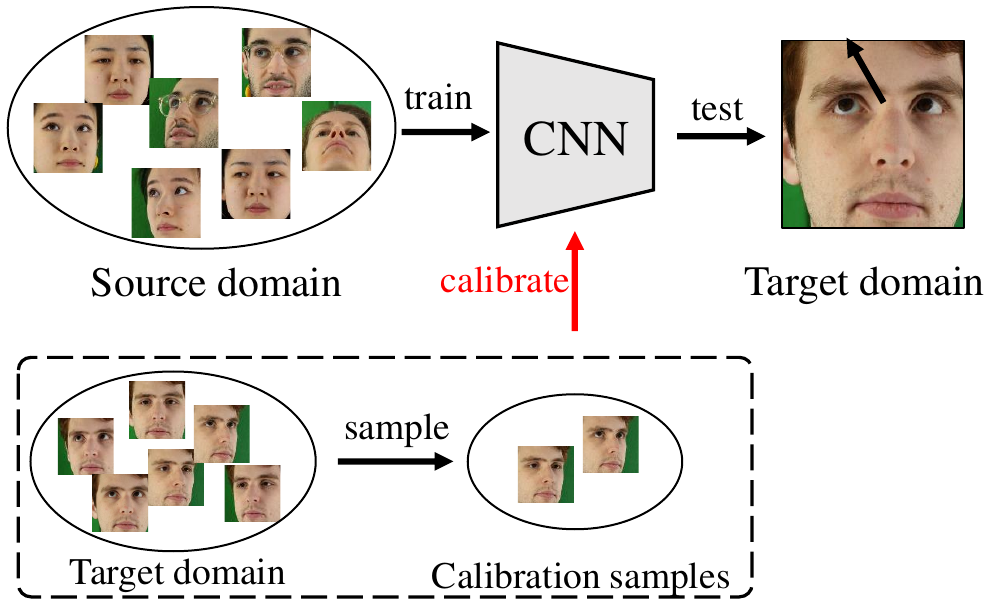} 
		\caption{Personal calibration in deep learning. The method usually samples a few images from the target domain as calibration samples. The calibration samples and training set are jointly used to improve the performance in target domain.}
            \vspace{-5mm}
		\label{fig:calibration}
	\end{figure}

	\subsection{Calibration}
	\label{ssec_calib}
	It is non-trivial to learn an accurate and universal gaze estimation model.	Conventional 3D eye model recovery methods usually build a unified gaze model including subject-specific parameters such as eyeball radius~\cite{Guestrin_2006_TBE}. They perform a personal calibration to estimate these subject-specific parameters. In the field of deep learning-based gaze estimation, 
	personal calibration is also explored to improve person-specific performance.
	~\fref{fig:calibration} shows a common pipeline of personal calibration in deep learning.

	
	
	\subsubsection{Calibration via Domain Adaptation}
	\label{sssec_dacalib}
	The calibration problem can be considered as domain adaption problems, where the training set is the source domain and the test set is the target domain. The test set usually contains unseen subjects or unseen environment. Researchers aim to improve the performance in the target domain using calibration samples. 
	
	The common approach of domain adaption is to fine-tune the model in the target domain~\cite{Krafka_2016_CVPR, Zhang_2018_CHI,li_2020_ICAICA}. 
	This is simple but effective. 
	Krafka \etal replace the fully-connected layer with an SVM and fine-tune the SVM layer to predict the gaze location~\cite{Krafka_2016_CVPR}. 
	Zhang~\etal split the CNN into three parts: the encoder, the feature extractor, and the decoder~\cite{Zhang_2018_CHI}. They fine-tune the encoder and decoder in each target domain. 
	Zhang~\etal also learn a third-order polynomial mapping function between the estimated and ground-truth of 2D gaze locations~\cite{Zhang_2019_CHI}. 
	Some studies introduce person-specific feature for gaze estimation~\cite{Lin_2019_iccvw,He_2019_iccvw}. 
	They learn the person-specific feature during fine-tuning. Linden~\etal introduce user embedding for recording personal information. 
	They obtain user embedding of the unseen subjects by fine-tuning using calibration samples~\cite{Lin_2019_iccvw}. Chen~\etal~\cite{Chen_2020_WACV,chen2022towards} observe the different gaze distributions of subjects. They use the calibration samples to estimate the bias between the estimated gaze and the ground-truth of different subjects. They use bias to refine the estimates.  Yu \etal generate additional calibration samples through the synthesis of gaze-redirected eye images from calibration samples~\cite{Yu_2019_CVPR}. The generated samples are also directly used for training.
	These methods all need labeled samples for supervised calibration. 
	
	\begin{figure}[t]
		\centering
		\includegraphics[width=0.9\columnwidth]{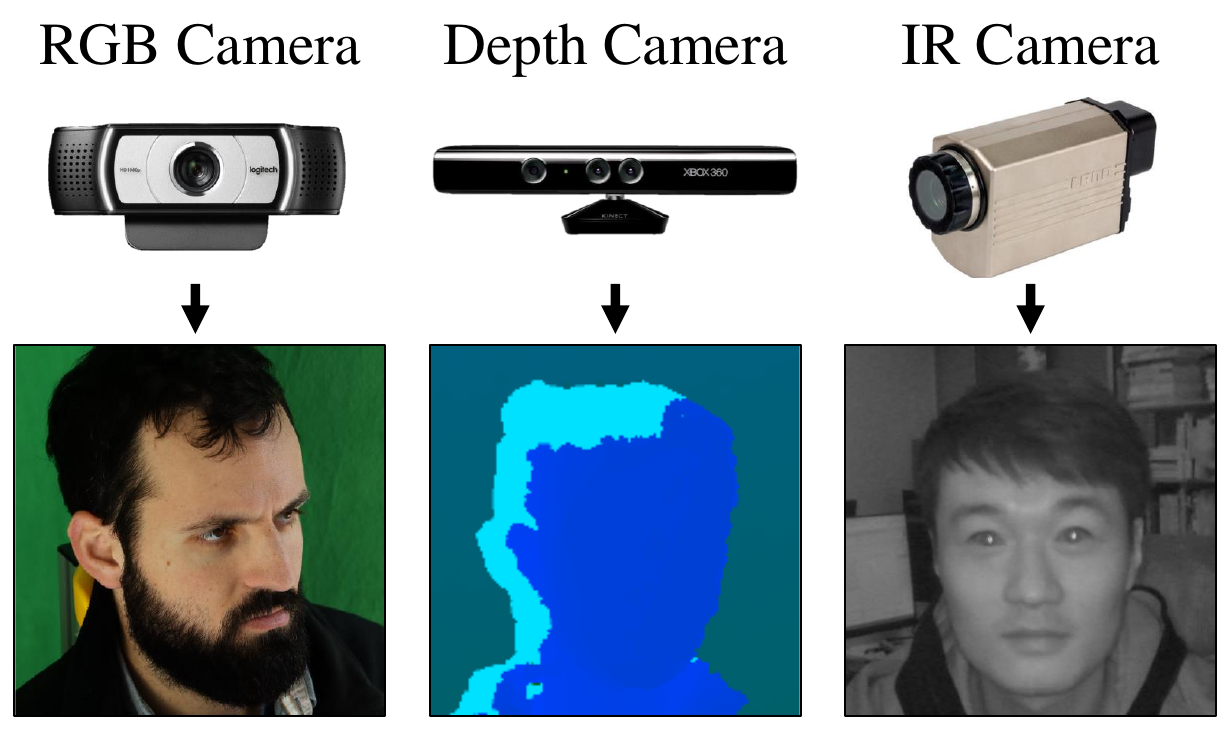} 
		\vspace{-2mm}
		\caption{Different cameras and their captured images.  }
		\label{fig:camera}
            \vspace{-5mm}
	\end{figure}
 
	Unsupervised calibration methods attract much attention recently. These methods use unlabeled calibration samples to improve performance.
	Wang~\etal propose an adversarial method for aligning features. They build a discriminator to judge the source of images from the extracted feature. The feature extractor has to confuse the discriminator,~\ie, the generated feature should be domain-invariant. Guo~\etal~\cite{guo_2020_ACCV} use source samples to form a locally linear representation of each target domain prediction in gaze space.
	The same linear relationships are applied in the feature space to generate the feature representation of target samples.
	Meanwhile, they minimize the difference between the generated feature and extracted feature of target sample for alignment.
	Cheng~\etal~\cite{cheng_2022_aaai} propose a domain generalization method. They improve the corss-dataset performance without knowing the target dataset or touching any new samples. They propose a self-adversarial framework to remove the gaze-irrelevant features in face images. Cui \etal define a new adaption problem~\cite{Cui_2017_ICIP}: adaptation from adults to children. 
	They use the conventional domain adaption method, geodesic flow kernel~\cite{Gong_2012_CVPR}, to transfer the feature in the adult domain into the children domain. 
	Bao \etal~\cite{bao2021story} estimate the point-of-regard by aligning the predicted gaze distribution with known gaze distribution.

    Some well-known strategies in universal tasks are proved effective for gaze estimation.
    Meta learning and metric learning show great potentials in personalized gaze estimation. They usually require few-shot annotated samples for calibration.
	Park \etal propose a meta learning-based calibration approach~\cite{Park_2019_ICCV}. They train a highly adaptable gaze estimation network through meta learning. 
	The network can be converted into a person-specific network once training with target person samples. 
	Liu \etal propose a differential CNN based on metric learning~\cite{Liu_2019_tpami}. 
	The network predicts gaze difference between two eye images. During test stage, it estimates the differences between inputs and calibration images, and takes the average results as the estimation.
 
    Contrastive learning and mean teacher~\cite{tarvainen2017mean} perform well in unsupervised domain adaption. They are usually used for cross-dataset task in gaze estimation. 
        Liu~\etal propose an outlier-guided collaborative learning for unsupervised cross-dataset tasks~\cite{liu2021generalizing}. 
        They create a group of teacher-student networks where teacher networks are pre-trained in source domain.
        They design the outlier-guided loss which requires the outputs of teacher and student networks to be consistent.
        Bao~\etal also propose a mean teacher architecture for unsupervised cross-dataset task~\cite{Bao_2022_CVPR}.
        They perform data augmentation \wrt~rotation in target domains and require the rotation consistency in gaze estimation.
        Wang~\etal \cite{wang2022contrastive} propose a contrastive learning for cross-dataset gaze estimation. They propose a contrastive loss function to encourage close feature distance for the samples with close gaze directions.


	\subsubsection{Calibration via User-unaware Data Collection}
	\label{sssec_unawarecalib}
	\added{It is difficult to acquire enough samples for calibration in practical applications. Collecting calibration samples in a user-unaware manner is an alternative solution~\cite{Chang_2019_iccvw, Wang_2016_etra,Klein_2019_BRACIS}.}
    Salvalaio~\etal implicitly collect calibration data when users are using computers. They collect data when the user is clicking a mouse, this is based on the assumption that users are gazing at the position of the cursor when clicking the mouse~\cite{Klein_2019_BRACIS}. They use online learning to fine-tune their model with the calibration samples.
	Some studies investigate the relation between the gaze points and the saliency maps~\cite{Kruthiventi_2016_CVPR, Wang_2020_tpami}. Chang \etal utilize saliency information to adapt the gaze cestimation algorithm to a new user without explicit calibration\cite{Chang_2019_iccvw}. They transform the saliency map into a differentiable loss map that can be used to optimize the CNN models. Wang \etal introduce a stochastic calibration procedure. They minimize the difference between the probability distribution of predicted gaze and ground truth~\cite{Wang_2016_etra}.
	\begin{figure}[t]
		\centering
		\includegraphics[width=1.0\columnwidth]{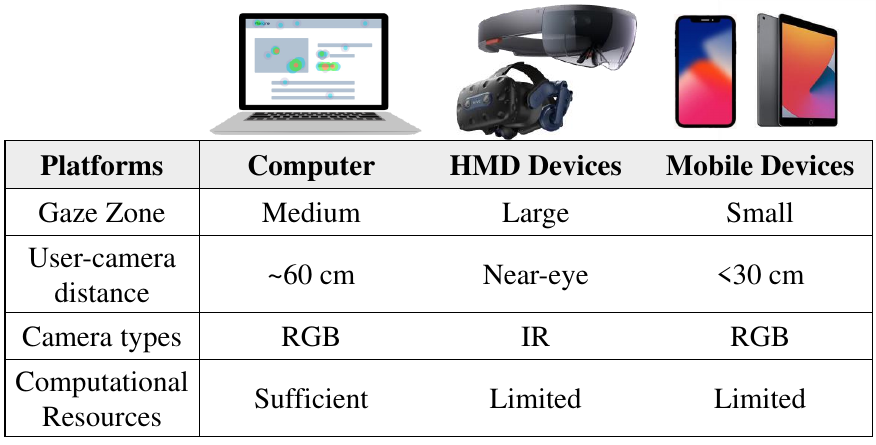} 
		\vspace{-5mm}
		\caption{Different platforms and their characteristics.}
        \vspace{-5mm}
		\label{fig:platform}
	\end{figure}

	\begin{table*}[!tp]
		\centering
		\caption{Summary of gaze estimation methods.}
            \vspace{-3mm}
		\label{tab:summary}
		\renewcommand\arraystretch{1.3}
		\setlength\tabcolsep{3pt}
		\begin{tabular}{c|c|c|c|c|c|c|c|c|c}
			\toprule[1.3pt]
			\multicolumn{2}{c|}{\multirow{2}{*}{\textbf{Perspectives}}}&\multicolumn{8}{c}{\textbf{Methods}}\\
			\cline{3-10}
			\multicolumn{2}{c|}{}&$2015$&$2016$&$2017$&$2018$&$2019$&$2020$&$2021$&$2022$\\
			\hline
	
			\multirow{6}{*}{\textbf{Feature}}&Eye
			image&\cite{Zhang_2015_CVPR}&---&\multicolumn{1}{m{1.0cm}|}{\cite{Zhang_2017_tpami,Tonsen_2017_IMWUT,Cui_2017_ICIP}}&\multicolumn{1}{m{1.5cm}<{\centering}|}{\cite{Park_2018_ECCV,Fischer_2018_ECCV,Cheng_2018_ECCV, Yu_2018_ECCVW,Lemley_2018_GEM}}&\multicolumn{1}{m{2.1cm}<{\centering}|}{\cite{Park_2019_ICCV,Chen_2019_ACCV,Wang2_2019_CVPR, Lian_2019_TNNLS,Wu_2019_iccvw,Liu_2019_tpami,Kim_2019_CHI}}&\multicolumn{1}{m{2.5cm}<{\centering}|}{\cite{Cheng_2020_AAAI, Bao_2020_ICPR,mahanama_2020_AH,kim_2020_ETRA,rangesh_2020_IV,lemley_2019_TCE, Yu_2020_CVPR}}&\multicolumn{1}{m{1.6cm}<{\centering}|}{\cite{biswas2021appearance,cai2021gaze,Sun_2021_ICCV}}&\multicolumn{1}{m{2.5cm}<{\centering}}{\cite{chen2022towards,bao2021story,li2022appearance,ghosh2022mtgls}}\\
			\cline{3-10}
			
			&Facial image&---&\cite{ Krafka_2016_CVPR}&\multicolumn{1}{m{1.0cm}<{\centering}|}{\cite{Zhang_2017_CVPRW,Deng_2017_ICCV,Recasens_2017_ICCV}}&\multicolumn{1}{m{1.5cm}<{\centering}|}{\cite{Palmero_2018_BMVC,Jyoti_2018_icpr,Chong_2018_ECCV,Zhang_2018_CHI}}&\multicolumn{1}{m{2.1cm}<{\centering}|}{\cite{Zhang_2019_CHI,Chen_2019_ACCV,Dubey_2019_IJCNN,Xiong_2019_CVPR,Lian_2019_AAAI,Lin_2019_iccvw,He_2019_iccvw,Chang_2019_iccvw,Klein_2019_BRACIS,Guo_2019_iccvw}}&\multicolumn{1}{m{2.5cm}<{\centering}|}{\cite{Cheng_2020_AAAI, Bao_2020_ICPR, Dias_2020_WACV, Zhang_2020_BMVC, Yu_2020_ICMI,zhuang_2021_IAEAC, zhang_2020_ICBSIC, zhao_2020_ICMA, liu_2020_iccpr, xia_2020_ICCSAE,wang_2020_WACV,guo_2020_ACCV, Zhang_2020_ECCV, Cheng_2020_tip, Chen_2020_WACV,mishra2020360}}&\multicolumn{1}{m{1.6cm}<{\centering}|}{\cite{kothari2021weakly,cai2021gaze,Li_2021_ICCV,liu2021generalizing,chen2020360}}&\multicolumn{1}{m{2.5cm}<{\centering}}{\cite{cheng2021gaze,zhang2022gazeonce,chen2022towards,Bao_2022_CVPR,wang2022contrastive,cheng_2022_aaai,Nonaka_2022_CVPR,lee2022latentgaze,kasahara2022look,yun2022haze,farkhondeh2022towards,O_Oh_2022_CVPR,qin2022learning,Gideon_2022_CVPR}}\\
			\cline{3-10}
			
			&Video&---&---&---&\cite{Palmero_2018_BMVC}&\multicolumn{1}{m{2.1cm}<{\centering}|}{\cite{Kellnhofer_2019_ICCV,Zhou_2019_ICME,Wang_2019_tvcg,Wang_2019_CVPR}}&\cite{park_2020_eccv}&\cite{kothari2021weakly}&---\\
			
			\midrule
		
			\multirow{7}{*}{\textbf{Model}}
			
			&\multicolumn{1}{p{0.11\textwidth}<{\centering}|}{Supervised CNN}&\cite{Zhang_2015_CVPR}&\cite{Krafka_2016_CVPR}&\multicolumn{1}{m{1.0cm}<{\centering}|}{\cite{Zhang_2017_CVPRW,Deng_2017_ICCV,Zhang_2017_tpami,Recasens_2017_ICCV,Cui_2017_ICIP,Tonsen_2017_IMWUT}}&\multicolumn{1}{m{1.5cm}<{\centering}|}{\cite{Park_2018_ECCV,Fischer_2018_ECCV,Palmero_2018_BMVC,Jyoti_2018_icpr,Yu_2018_ECCVW,Chong_2018_ECCV,Zhang_2018_CHI,Lemley_2018_GEM}}&\multicolumn{1}{m{2.1cm}<{\centering}|}{\cite{Zhang_2019_CHI,Kellnhofer_2019_ICCV,Park_2019_ICCV,Chen_2019_ACCV,Zhou_2019_ICME,Wang_2019_tvcg,Wang_2019_CVPR,Lian_2019_TNNLS,Lian_2019_AAAI,Wu_2019_iccvw,Liu_2019_tpami,Chang_2019_iccvw,Klein_2019_BRACIS,Kim_2019_CHI,Guo_2019_iccvw}}&\multicolumn{1}{m{2.5cm}<{\centering}|}{\cite{Zhang_2020_ECCV,Bao_2020_ICPR,Cheng_2020_AAAI,Cheng_2020_tip,Zhang_2020_BMVC,Yu_2020_ICMI,zhuang_2021_IAEAC,zhang_2020_ICBSIC,zhao_2020_ICMA,liu_2020_iccpr,mahanama_2020_AH,xia_2020_ICCSAE,kim_2020_ETRA,rangesh_2020_IV,lemley_2019_TCE,wang_2020_WACV,park_2020_eccv,Chen_2020_WACV,mishra2020360}}&\multicolumn{1}{m{1.6cm}<{\centering}|}{\cite{biswas2021appearance,cai2021gaze,Li_2021_ICCV,chen2020360}}&\multicolumn{1}{m{2.5cm}<{\centering}}{\cite{cheng2021gaze,zhang2022gazeonce,chen2022towards,cheng_2022_aaai,Nonaka_2022_CVPR,li2022appearance,kasahara2022look,yun2022haze,O_Oh_2022_CVPR}}\\
			\cline{3-10}
			
			&\multicolumn{1}{p{0.11\textwidth}<{\centering}|}{Semi-/Self-/Un- Supervised CNN}&---&---&---&\cite{Cheng_2018_ECCV}&\multicolumn{1}{m{2.1cm}<{\centering}|}{\cite{Wang2_2019_CVPR,Dubey_2019_IJCNN,Xiong_2019_CVPR, Lin_2019_iccvw,He_2019_iccvw}}&\cite{Yu_2020_CVPR,guo_2020_ACCV,Cheng_2020_tip}&\multicolumn{1}{m{1.6cm}<{\centering}|}{\cite{kothari2021weakly,bao2021story,Sun_2021_ICCV,liu2021generalizing}}&\multicolumn{1}{m{2.5cm}<{\centering}}{\cite{zhang2022gazeonce,Bao_2022_CVPR,wang2022contrastive,cheng_2022_aaai,lee2022latentgaze,bao2021story,ghosh2022mtgls,farkhondeh2022towards,Gideon_2022_CVPR,qin2022learning}}\\
			\cline{3-10}
			
			&\multicolumn{1}{p{0.11\textwidth}<{\centering}|}{Multi-task CNN}&---&---&\cite{Recasens_2017_ICCV, Deng_2017_ICCV}&\cite{Yu_2018_ECCVW,Chong_2018_ECCV}&\multicolumn{1}{m{2.1cm}<{\centering}|}{\cite{Kellnhofer_2019_ICCV,Lian_2019_TNNLS,Lian_2019_AAAI,Wu_2019_iccvw}}&\cite{park_2020_eccv}&---&\cite{ghosh2022mtgls}\\
			\cline{3-10}
			
			&\multicolumn{1}{p{0.11\textwidth}<{\centering}|}{Recurrent CNN}&---&---&---&\cite{Palmero_2018_BMVC}&\cite{Kellnhofer_2019_ICCV,Zhou_2019_ICME}&\cite{Cheng_2020_AAAI,park_2020_eccv}&---&---\\
			\cline{3-10}
			
			&\multicolumn{1}{p{0.11\textwidth}<{\centering}|}{CNN with Priors}&---&---&\cite{Deng_2017_ICCV}&\cite{Park_2018_ECCV, Cheng_2018_ECCV}&\cite{Wang_2019_CVPR, Xiong_2019_CVPR}&{\cite{Chen_2020_WACV}}&---&\cite{chen2022towards}\\
			
			\midrule
			\multirow{3}{*}{\textbf{Calibration}} & \multicolumn{1}{p{0.11\textwidth}<{\centering}|}{Domain Adaption}&---&\cite{Krafka_2016_CVPR}&\cite{Cui_2017_ICIP}&\cite{ Zhang_2018_CHI}&\multicolumn{1}{m{2.1cm}<{\centering}|}{\cite{Zhang_2019_CHI,Yu_2019_CVPR,Park_2019_ICCV,Lin_2019_iccvw,He_2019_iccvw,Liu_2019_tpami}}&\multicolumn{1}{m{2.5cm}<{\centering}|}{\cite{Chen_2020_WACV,li_2020_ICAICA,guo_2020_ACCV}}&\cite{bao2021story,liu2021generalizing}&\multicolumn{1}{m{2.5cm}<{\centering}}{\cite{chen2022towards,Bao_2022_CVPR,wang2022contrastive,lee2022latentgaze,bao2021story,ghosh2022mtgls,farkhondeh2022towards}}\\
			\cline{3-10}
			
			& \multicolumn{1}{p{0.11\textwidth}<{\centering}|}{User-unaware Data Collection}&\multirow{2}{*}{---}&\multirow{2}{*}{\cite{Wang_2016_etra}}&\multirow{2}{*}{---}&\multirow{2}{*}{---}&\multirow{2}{*}{\cite{Chang_2019_iccvw, Klein_2019_BRACIS}}&\multirow{2}{*}{---}&\multirow{2}{*}{---}&\multirow{2}{*}{---}\\
			
		
			\midrule
			\multirow{5}{*}{\textbf{Camera}}&\multicolumn{1}{p{0.11\textwidth}<{\centering}|}{Single camera} &\cite{Zhang_2015_CVPR}&\cite{Krafka_2016_CVPR}&\multicolumn{1}{m{1.0cm}<{\centering}|}{\cite{Zhang_2017_CVPRW,Deng_2017_ICCV,Zhang_2017_tpami,Cui_2017_ICIP}}&\multicolumn{1}{m{1.5cm}<{\centering}|}{\cite{Park_2018_ECCV,Fischer_2018_ECCV,Cheng_2018_ECCV,Palmero_2018_BMVC,Jyoti_2018_icpr,Yu_2018_ECCVW,Chong_2018_ECCV,Zhang_2018_CHI,Lemley_2018_GEM}}&\multicolumn{1}{m{2.1cm}<{\centering}|}{\cite{Zhang_2019_CHI,Kellnhofer_2019_ICCV,Park_2019_ICCV,Chen_2019_ACCV,Dubey_2019_IJCNN,Zhou_2019_ICME,Wang_2019_tvcg,Wang_2019_CVPR,Wu_2019_iccvw,He_2019_iccvw,Liu_2019_tpami,Chang_2019_iccvw,Klein_2019_BRACIS,Kim_2019_CHI,Guo_2019_iccvw}}&\multicolumn{1}{m{2.5cm}<{\centering}|}{\cite{Zhang_2020_BMVC,Cheng_2020_AAAI,Bao_2020_ICPR,Yu_2020_ICMI,zhuang_2021_IAEAC,zhang_2020_ICBSIC,zhao_2020_ICMA,liu_2020_iccpr,mahanama_2020_AH,xia_2020_ICCSAE,kim_2020_ETRA,rangesh_2020_IV,lemley_2019_TCE,wang_2020_WACV,park_2020_eccv,Zhang_2020_ECCV,Cheng_2020_tip,Chen_2020_WACV,mishra2020360}}&\multicolumn{1}{m{1.6cm}<{\centering}|}{\cite{biswas2021appearance,bao2021story,cai2021gaze,kothari2021weakly,Sun_2021_ICCV,liu2021generalizing,chen2020360}}&\multicolumn{1}{m{2.5cm}<{\centering}}{\cite{cheng2021gaze,zhang2022gazeonce,chen2022towards,Bao_2022_CVPR,wang2022contrastive,cheng_2022_aaai,Nonaka_2022_CVPR,lee2022latentgaze,bao2021story,li2022appearance,kasahara2022look,ghosh2022mtgls,yun2022haze,farkhondeh2022towards,O_Oh_2022_CVPR,qin2022learning}}\\
			\cline{3-10}
			
			& \multicolumn{1}{p{0.11\textwidth}<{\centering}|}{Multi cameras} &---&---&\cite{Tonsen_2017_IMWUT}&---&\cite{Lian_2019_TNNLS}&---&---&\cite{Gideon_2022_CVPR}\\
			\cline{3-10}
			
			& \multicolumn{1}{p{0.11\textwidth}<{\centering}|}{IR Camera}&---&---&---&---&\cite{Wu_2019_iccvw, Kim_2019_CHI}&\cite{rangesh_2020_IV}&---&---\\
			\cline{3-10}
			
			&\multicolumn{1}{p{0.11\textwidth}<{\centering}|}{RGBD Camera}&---&---&---&---&\cite{Lian_2019_AAAI}&---&---&---\\
			\cline{3-10}
			
			&\multicolumn{1}{p{0.11\textwidth}<{\centering}|}{Near-eye Camera}&---&---&\cite{Tonsen_2017_IMWUT}&---&\cite{Wu_2019_iccvw,Kim_2019_CHI}&---&---&---\\
			
			\midrule
		
			\multirow{4}{*}{\textbf{Platform}} & Computer &\cite{Zhang_2015_CVPR} &---&\multicolumn{1}{m{1.0cm}<{\centering}|}{\cite{Zhang_2017_CVPRW,Deng_2017_ICCV,Zhang_2017_tpami,Cui_2017_ICIP}}&\multicolumn{1}{m{1.5cm}<{\centering}|}{\cite{Park_2018_ECCV,Fischer_2018_ECCV,Cheng_2018_ECCV,Palmero_2018_BMVC,Jyoti_2018_icpr,Yu_2018_ECCVW,Chong_2018_ECCV,Zhang_2018_CHI}}&\multicolumn{1}{m{2.1cm}<{\centering}|}{\cite{Zhang_2019_CHI,Kellnhofer_2019_ICCV,Park_2019_ICCV,Chen_2019_ACCV,Dubey_2019_IJCNN,Zhou_2019_ICME,Wang_2019_tvcg,Wang_2019_CVPR,Lian_2019_TNNLS,Lian_2019_AAAI,Liu_2019_tpami,Chang_2019_iccvw,Klein_2019_BRACIS}}&\multicolumn{1}{m{2.5cm}<{\centering}|}{\cite{Cheng_2020_AAAI,Zhang_2020_BMVC,Yu_2020_ICMI,zhuang_2021_IAEAC,zhang_2020_ICBSIC,zhao_2020_ICMA,liu_2020_iccpr,mahanama_2020_AH,kim_2020_ETRA,rangesh_2020_IV,lemley_2019_TCE,wang_2020_WACV,park_2020_eccv,Zhang_2020_ECCV,Cheng_2020_tip,Chen_2020_WACV,mishra2020360}}&\multicolumn{1}{m{1.6cm}<{\centering}|}{\cite{biswas2021appearance,bao2021story,cai2021gaze,kothari2021weakly,Sun_2021_ICCV,liu2021generalizing,chen2020360}}&\multicolumn{1}{m{2.5cm}<{\centering}}{\cite{cheng2021gaze,zhang2022gazeonce,chen2022towards,Bao_2022_CVPR,wang2022contrastive,cheng_2022_aaai,Nonaka_2022_CVPR,lee2022latentgaze,bao2021story,li2022appearance,ghosh2022mtgls,yun2022haze,farkhondeh2022towards,Gideon_2022_CVPR,O_Oh_2022_CVPR,qin2022learning}}\\
			\cline{3-10}
			
			&\multicolumn{1}{p{0.11\textwidth}<{\centering}|}{Mobile Device}&---&\cite{Krafka_2016_CVPR}&---&\cite{Zhang_2018_CHI}&\cite{He_2019_iccvw,Guo_2019_iccvw}&\cite{ Bao_2020_ICPR,xia_2020_ICCSAE}&---&---\\
			\cline{3-10}
			
			&\multicolumn{1}{p{0.11\textwidth}<{\centering}|}{HMD Device}&---&---&\cite{Tonsen_2017_IMWUT}&\cite{Lemley_2018_GEM}&\cite{Wu_2019_iccvw,Kim_2019_CHI}&---&---&---\\
			\bottomrule[1.3pt]
		\end{tabular}
	\end{table*}

	\subsection{Devices and Platforms}
	\label{ssec_device}
	
	\subsubsection{Camera}
	\label{sssec_camera}
	
	\added{The majority of gaze estimation systems use a single RGB camera to capture eye images, while some studies use different camera settings, \eg, using multiple cameras to capture multi-view images~\cite{Lian_2019_TNNLS, Tonsen_2017_IMWUT,Cheng_2023_ICCV}, using infrared (IR) cameras to handle low illumination condition~\cite{Wu_2019_iccvw, Kim_2019_CHI}, and using RGBD cameras to provide the depth information~\cite{Lian_2019_AAAI}.} Different cameras and their captured images are shown in~\fref{fig:camera}.
 
	Tonsen~\etal embed multiple millimeter-sized RGB cameras into a normal glasses frame~\cite{Tonsen_2017_IMWUT}. They use multi-layer perceptrons to process the eye images captured by different cameras, and concatenate the extracted feature to estimate gaze.	
	Lian~\etal mount three cameras at the bottom of a screen~\cite{Lian_2019_TNNLS}. 
	They build a multi-branch network to extract the features of each view and concatenate them to estimate $2$D gaze position on the screen. Wu~\etal collect gaze data using near-eye IR cameras~\cite{Wu_2019_iccvw}. They use CNN to detect the location of glints, pupil centers and corneas from IR images. Then, they build an eye model using the detected feature and estimate gaze from the gaze model. Kim~\etal collect a large-scale dataset of near-eye IR eye images~\cite{Kim_2019_CHI}. They synthesize additional IR eye images that cover large variations in face shape, gaze direction, pupil and iris~\etc. Lian \etal use RGBD cameras to capture depth facial images~\cite{Lian_2019_AAAI}. They extract the depth information of eye regions and concatenate it with RGB image features to estimate gaze. 
	
	
	\begin{figure}[t]
		\centering
		\includegraphics[width=\columnwidth]{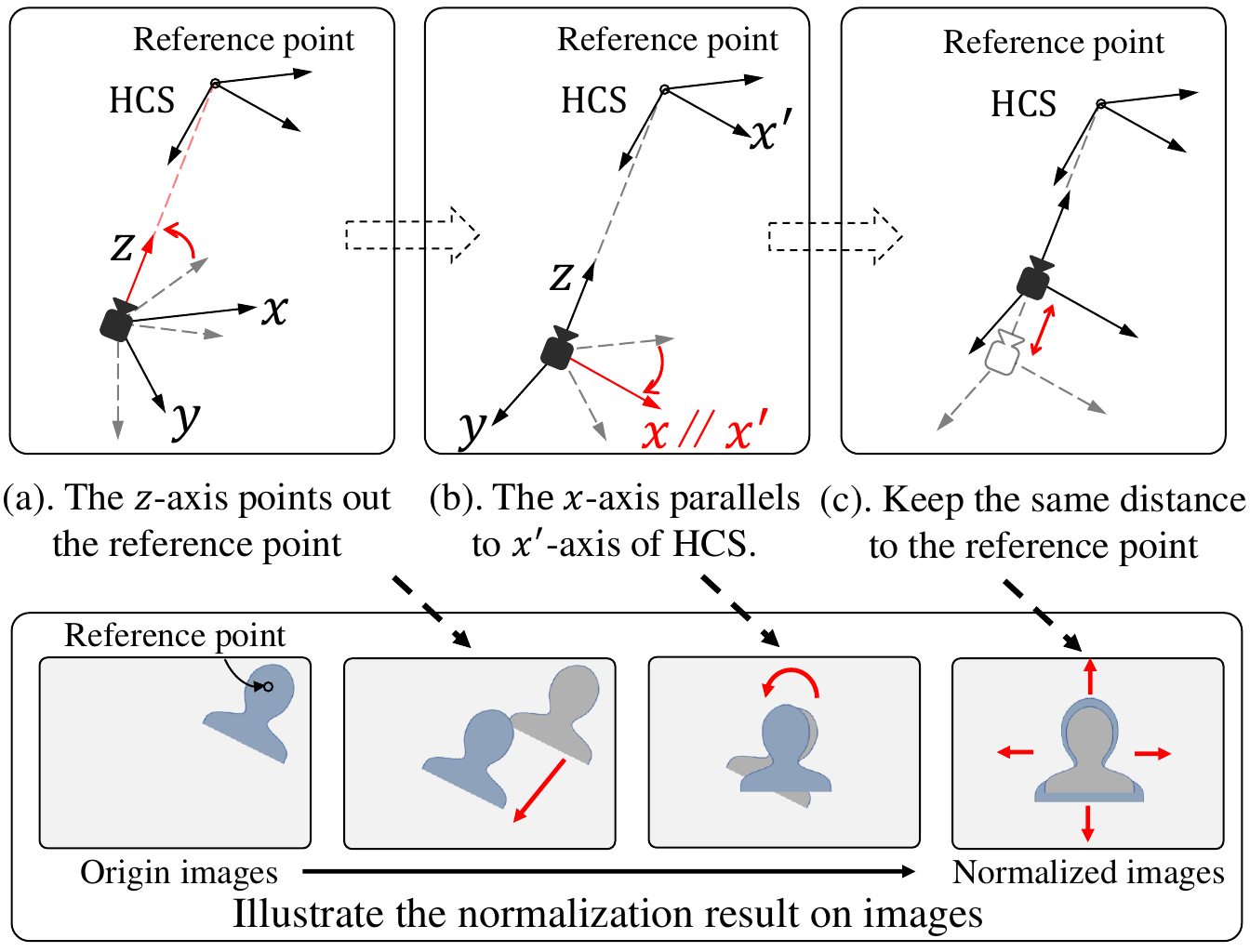} 
		\caption{A data rectification method~\cite{Sugano_2014_CVPR}. The virtual camera is rotated so that the $z$-axis points at the reference point and the $x$-axis is parallel with the $x^\prime$-axis of the head coordinate system (HCS).
			The bottom row illustrates the rectification result on images. Overall, the reference point is moved to the center of images, the image is rotated to straighten face and scaled to align the size of face in different images.  }
        \vspace{-5mm}
		\label{NormFig}
	\end{figure}
 
	\subsubsection{Platform}
	\label{sssec_platform}
	Eye gaze can be used to estimate human intent in various applications,~\eg,~product design evaluation~\cite{Khalighy_2015_IE}, marketing studies~\cite{Santos_2015_IJPS} and human-computer interaction~\cite{Zhang_2017_uist, Sugano_2016_uist,Wang_2015_Hybrid}. 
	These applications can be simply categorized into three types of platforms: computers, mobile devices and head-mounted devices. We summarize the characteristics of these platforms in~\fref{fig:platform}.
	
	The computer is the most typical platform for appearance-based gaze estimation. The cameras are usually placed below/above the computer screen~\cite{Zhang_2015_CVPR, Zhang_2016_tpami, Cheng_2018_ECCV, Cheng_2020_AAAI, Park_2018_ECCV}. Some works focus on using deeper neural networks~\cite{Zhang_2015_CVPR,Zhang_2017_CVPRW,Fischer_2018_ECCV} or extra modules~\cite{Park_2018_ECCV,Cheng_2018_ECCV,Cheng_2020_AAAI} to improve gaze estimation performance, while the other studies seek to use custom devices for gaze estimation, such as multi-cameras and RGBD cameras~\cite{Lian_2019_TNNLS,Lian_2019_AAAI}.

	The mobile device contains front cameras but has limited computational resources. The related methods usually estimate PoG instead of gaze directions due to the difficulty of geometric calibration. Krafka \etal propose iTracker for mobile devices~\cite{Krafka_2016_CVPR}, which combines the facial image, two eye images and the face grid to estimate the gaze. The face grid encodes the position of the face in captured images and is proved to be effective for gaze estimation in mobile devices in many works~\cite{He_2019_iccvw, Bao_2020_ICPR}. He~\etal propose a more accurate and faster method based on iTracker~\cite{He_2019_iccvw}. They replace the face grid with eye corner landmark feature. Guo~\etal propose a generalized gaze estimation method~\cite{Guo_2019_iccvw}. They observe the notable jittering problem in gaze point estimates and propose to use adversarial training to address this problem.
	Valliappan~\cite{valliappan2020accelerating} evaluate the eye tracking with deep learning on smartphone.
	They show the algorithm can achieve competitive result compared with modern eye tracking devices.
	
	The head-mounted device usually employs near-eye cameras to capture eye images.	Tonsen \etal embed millimetre-sized RGB cameras into a normal glasses frame~\cite{Tonsen_2017_IMWUT}. In order to compensate for the low-resolution captured images, they use multi-cameras to capture multi-view images and use a neural network to regress gaze from these images.	
	IR cameras are also employed by head-mounted devices. 
	Wu \etal collect the MagicEyes dataset using IR cameras \cite{Wu_2019_iccvw}. They propose EyeNet, a neural network that solves multiple heterogeneous tasks related to eye gaze estimation for an off-axis camera setting. They use the CNN to model $3$D cornea and $3$D pupil and estimate the gaze from these two $3$D models. Lemley \etal use the single near-eye image as input to the neural network and directly regress gaze \cite{Lemley_2018_GEM}. Kim \etal follow a similar approach and collect the NVGaze dataset \cite{Kim_2019_CHI}.
	
	
	\subsection{Summary}
	\label{ssec_summarization}
	
	\Tref{tab:summary} summarizes the existing CNN-based gaze estimation methods. Note that many methods do not specify a platform~\cite{Zhang_2015_CVPR,Cheng_2020_AAAI}. Thus, we categorize these methods into the platform of "computer". In general, there is an increasing trend in developing supervised or semi-/self-/un-supervised CNN structures to estimate gaze. Many recent research interests shift to different calibration approaches through domain adaptation or user-unaware data collection. The first CNN-based gaze direction estimation method is proposed by Zhang \etal in 2015~\cite{Zhang_2015_CVPR}, the first CNN-based PoG estimation method is proposed by Krafka \etal in 2016~\cite{Krafka_2016_CVPR}. These two studies both provide large-scale gaze datasets, the MPIIGaze and the GazeCapture, which have been widely used for evaluating gaze estimation algorithms in later studies.
	

	\begin{table}[tb]
		\centering
		\caption{Summary of face alignment methods}
        \vspace{-3mm}
		\label{tab:detection}
		\renewcommand\arraystretch{1.1}
		\setlength\tabcolsep{2pt}
		\begin{tabular}{p{2.0cm}|c|l|p{4.8cm}}
			\toprule[1.3pt]
			\textbf{Names}&\textbf{Years}& \textbf{Pub.} &\textbf{Links}\\
		
			\hline
			Dlib~\cite{Kazemi_2014_CVPR}&2014&CVPR&\url{https://pypi.org/project/dlib/19.6.0/}\\
			MTCNN~\cite{Zhang_2016_SPL} &2016&SPL&\url{https://github.com/kpzhang93/MTCNN_face_detection_alignment}\\
			DAN~\cite{Kowalski_2017_CVPRW}&2017&CVPRW&\url{https://github.com/MarekKowalski/DeepAlignmentNetwork}\\
			OpenFace~\cite{Baltrusaitis_2018_FG} &2018&FG & \url{https://github.com/TadasBaltrusaitis/OpenFace}\\
			PRN~\cite{Feng_2018_ECCV}&2018&ECCV&\url{https://github.com/YadiraF/PRNet}\\
			3DDFA\_V2~\cite{Guo_2020_ECCV}	&2020	&ECCV &\url{https://github.com/cleardusk/3DDFA_V2}\\
			\bottomrule[1.3pt]
			
		\end{tabular}
        \vspace{-5mm}
	\end{table}

	\section{Datasets and Benchmarks}
	\label{sec_dataset}
	
	\subsection{Data Pre-processing}
	\label{ssec_datapre}

	\subsubsection{Face and Eye Detection}
	\label{sssec_detect}
	Raw images often contain unnecessary information for gaze estimation, such as the background. 
	Directly using raw images to regress gaze not only increases the computational resource but also brings nuisance factors such as changes in scenes.
	Therefore, face or eye detection is usually applied in raw images to prune unnecessary information. 
	Generally, researchers first perform face alignment in raw images to obtain facial landmarks and crop face/eye images using these landmarks. 
	Several face alignment methods have been proposed recently~\cite{Zhang_2020_TIP, Chandran_2020_CVPR, Gao_2020_TM}. 
	We list some typical face alignment methods in~\Tref{tab:detection}.
	
	After the facial landmarks are obtained, face or eye image are cropped accordingly. 
	There is no protocol to regulate the cropping procedure. 
	We provide a common cropping procedure here as an example.	
	We let $x_i\in \mathbb{R}^2$ be the x, y-coordinates of the $i$th facial landmark in an raw image $I$.	
	The center point $\overline{x}$ is calculated as $\overline{x}=\frac{1}{n}\sum_{i=1}^{n}x_i$, where $n$ is the number of facial landmarks. The face image is defined as a square region with the center $\overline{x}$ and an width $w$. 
	The $w$ is usually set empirically. 
	For example, ~\cite{Zhang_2017_CVPRW} set $w$ as $1.5$ times of the maximum distance between the landmarks. 
	The eye cropping is similar to face cropping, while the eye region is usually defined as a rectangle with the center set as the centroid of eye landmarks.	
	The width of the rectangle is set based the distance between eye corners,~\eg, 1.2 times.

	\subsubsection{Data Rectification}
	\label{sssec_rectify}
	


    Data rectification eliminate environment factors such as head pose and illumination. It simplifies gaze regression problem with data pre-processing methods. 
    Sugano~\etal propose to rectify the eye image by rotating the virtual camera to point at the same reference point in the human face \cite{Sugano_2014_CVPR}. 
	They assume that the captured eye image is a plane in 3D space, the rotation of the virtual camera can be performed as a perspective transformation on the image. 
	The whole data rectification process is shown in ~\Fref{NormFig}. 
	They compute the transformation matrix $\bm{M} = \bm{SR}$, where $\bm{R}$ is the rotation matrix and $\bm{S}$ is the scale matrix. $\bm{R}$ also indicates the rotated camera coordinate system. 
	The $z$-axis $\bm{z_c}$ of the rotated camera coordinate system is defined as the line from cameras to reference points, where the reference point is usually set as the face center or eye center. 
	It means that the rotated camera is pointing towards the reference point. 
	The rotated $x$-axis $\bm{x_c}$ is defined as the $x$-axis of the head coordinate system so that the appearance captured by the rotated cameras is facing the front. 
	The rotated $y$-axis $\bm{y_c}$ can be computed by $\bm{y_c}= \bm{z_c} \times \bm{x_c}$, the $\bm{x_c}$ is recalculated by $\bm{x_c} = \bm{y_c} \times \bm{z_c}$ to maintain orthogonality. 
	As a result, the rotation matrix $\bm{R}=[\frac{\bm{x_c}}{||\bm{x_c}||}, \frac{\bm{y_c}}{||\bm{y_c}||}, \frac{\bm{z_c}}{||\bm{z_c}||}]$. 
	The $\bm{S}$ maintains the distance between the virtual camera and the reference point, which is defined as $diag(1, 1, \frac{d_n}{d_o})$, where $d_o$ is the original distance between the camera and the reference point, and $d_n$ is the new distance that can be adjusted manually. 
	They apply a perspective transformation on images with $\bm{W} = \bm{C_nMC_r^{-1}}$, where $\bm{C_r}$ is the intrinsic matrix of the original camera and $\bm{C_n}$ is the intrinsic matrix of the new camera. 
	Gaze directions can also be calculated in the rotated camera coordinate system as $\hat{\bm{g}} = \bm{Mg}$. The method eliminates the ambiguity caused by different head positions and aligns the intrinsic matrix of cameras. It also rotates the captured image to cancel the degree of freedom of roll in head rotation. Zhang \etal further explore the method in ~\cite{Zhang_2018_etra}. They argue that scaling can not change the gaze direction vector. The gaze direction is computed by $\hat{\bm{g}} = \bm{Rg}$.
	
	Illumination also influences the appearance of the human eye. 
	To handle this, researchers usually take gray-scale images rather than RGB images as input and apply histogram equalization in the gray-scale images to enhance the image.

\begin{table}[tb]
	\centering
	\caption{Symbol table in data post-processing}
	\vspace{-3mm}
	\label{tab:symbol}
	\renewcommand\arraystretch{1.4}
	\setlength\tabcolsep{1pt}
	\begin{tabular}{p{43pt}p{200pt}}
		\toprule[1.3pt]
		\textbf{Symbol}& \textbf{Meaning} \\
		
		\hline
		$\bm{p}\in \mathbb{R}^2$ & $\bm{p}=(u, v)$,  gaze targets.\\
		$\bm{g}\in \mathbb{R}^3$ & $\bm{g}=(g_x, g_y, g_z)$, gaze directions. \\
		$\bm{o}\in\mathbb{R}^3$ & $\bm{o}=(x_o, y_o, z_o)$, origins of gaze directions.\\
		$\bm{t}\in\mathbb{R}^3$ & $\bm{t}=(x_t, y_t, z_t)$, targets of gaze directions.\\
		$\bm{R_s}\in\mathbb{R}^{3\times3}$ & The rotation matrix of SCS \wrt~CCS. \\
		$\bm{T_s}\in\mathbb{R}^3$ & $\bm{T_s}=(t_x, t_y, t_z)$, the translation matrix of SCS \wrt~CCS. \\
		$\bm{n}\in\mathbb{R}^3$ & $\bm{n}=(n_x, n_y, n_z)$, the normal vectors of x-y plane of SCS.\\
		\bottomrule[1.3pt]
		
	\end{tabular}
	\vspace{-5mm}
\end{table}
	\subsection{Data Post-processing}
	\label{ssec_datapost}
\begin{figure}[t]
		\centering
		\includegraphics[width=0.9\columnwidth]{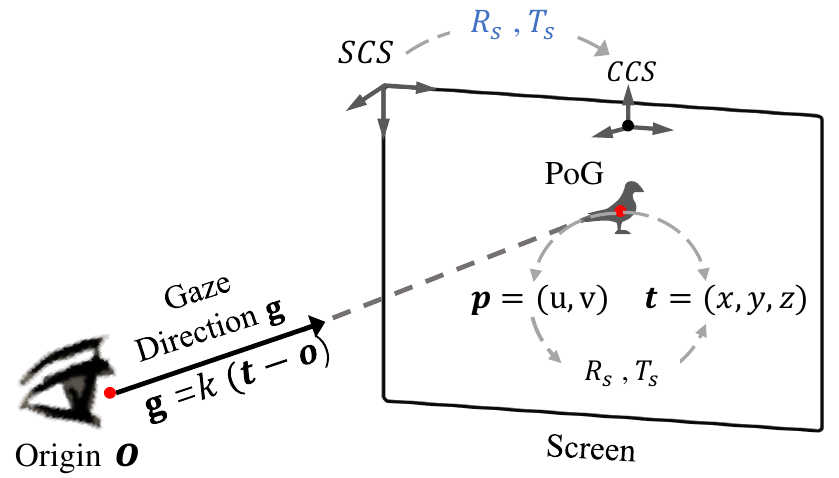} 
		\caption{We illustrate the relation between gaze directions and PoG. Gaze directions are originated from an origin $\bm{o}$ and intersect with the screen at the PoG. The PoG is usually denoted as a 2D coordinate $\bm{p}$. \added{It can be converted to 3D coordinate $\bm{t}$ in CCS with screen pose $\{R_s,T_s \}$. Gaze directions can be computed with $k(\bm{t}-\bm{o})$, where $k$ is a scale factor.}}
        \vspace{-5mm}
		\label{SymbolFig}
	\end{figure}

	Various applications require different forms of gaze estimates. 
	For example, in a real-world interaction task, it requires 3D gaze direction to estimate the human intent \cite{Wang_2019_Gaze,Wang_2018_Slam}, while it requires 2D PoG for the screen-based interaction \cite{Wang_2015_Hybrid,Dong_2015_Hybrid}. In this section, we introduce how to convert different forms of gaze estimates by post-processing. 
	We list the symbols in \Tref{tab:symbol} and illustrate the symbols in~\fref{SymbolFig}. 
	We denote the PoG as 2D gaze and the gaze direction as 3D gaze in this section.

	\subsubsection{2D/3D Gaze Conversion}
	\label{sssec_2d3d}
	The 2D gaze estimation algorithm usually estimates gaze targets on a computer screen~\cite{Guo_2019_iccvw, Wang_2019_CVPR, Krafka_2016_CVPR, Chang_2019_iccvw, Wong_2019_Percom}, while the 3D gaze estimation algorithm estimates gaze directions in 3D space~\cite{Zhang_2017_tpami, Zhang_2017_CVPRW, Kellnhofer_2019_ICCV, Cheng_2020_AAAI, Xiong_2019_CVPR}. 
	We first introduce how to convert between the 2D gaze and the 3D gaze.
	
	Given a 2D gaze target $\bm{p}=(u, v)$ on the screen, our goal is to compute the corresponding 3D gaze direction $\bm{g}=(g_x, g_y, g_z)$.	
	The processing pipeline is that we first compute the 3D gaze target $\bm{t}$ and 3D gaze origin $\bm{o}$ in the camera coordinate system (CCS). 
	The gaze direction can be computed as 
	\begin{equation}
		\bm{g} = \frac{\bm{t}-\bm{o}}{||\bm{t}-\bm{o}||}.
	\end{equation}
	
	\begin{table*}[htp]
		\setlength\tabcolsep{5pt}
		\renewcommand\arraystretch{1.3}
		\renewcommand{\multirowsetup}{\centering}
		\normalsize
		\caption{Summary of common gaze estimation datasets.}
        \vspace{-3mm}
		\begin{threeparttable}
			\begin{tabular}{p{3.7cm}|c|c|c|c|c|p{4.8cm}|p{2cm}}
				\toprule
				
				\multirow{2}{*}{\textbf{Datasets}} & \multirow{2}{*}{\textbf{Subjects}} & \multirow{2}{*}{\textbf{Total}} & \multicolumn{3}{c|}{\textbf{Annotations}} &\multirow{2}{*}{\textbf{Brief Introduction}}  &\multirow{2}{*}{\textbf{Links}}\\
				\cline{4-6}
				&&& \makecell{Full\\face} &\makecell{$2$D\\Gaze} &\makecell{$3$D\\Gaze}&&\\
				\hline
				\hline
				Columbia~\cite{Smith_2013_UIST}, 2013, \textcolor{white}{**} {\small(Columbia University)}	& $58$ & $6$K images & \checkmark &$\times$&\checkmark & Collected in laboratory; 5 head pose and 21 gaze directions per head pose.	&\small{\url{https://cs.columbia.edu/CAVE/databases/columbia_gaze}} \\
				\hline
				UTMultiview~\cite{Sugano_2014_CVPR}, 2014,   {\small(The University of Tokyo; Microsoft Research Asia)}	& $50$ & $1.1$M images 	&$\times$&\checkmark&\checkmark	&Collected in laboratory; Fixed head pose; Multiview eye images; 
				Synthesis eye images. &\small{\url{https://ut-vision.org/datasets}} \\
				\hline
				EyeDiap~\cite{Mora_2014_ETRA}, 2014, \textcolor{white}{***} {\small(Idiap Research Institute)}   	   &$16$  & $94$ videos     &\checkmark&  \checkmark &\checkmark  &Collected in laboratory; Free head poes;  Additional depth videos. &\small{\url{https://idiap.ch/dataset/eyediap}}  \\
				\hline
				MPIIGaze~\cite{Zhang_2017_tpami}, 2015, \textcolor{white}{***} {\small(Max Planck Institute)} 	  & $15$ & $213$K images&$\times$ & \checkmark &\checkmark  & Collected by laptops in daily life; Free head pose and illumination. &\small{\url{https://mpi-inf.mpg.de/mpiigaze}}\\
				\hline
				GazeCapture~\cite{Krafka_2016_CVPR}, 2016,\textcolor{white}{*} {\small(University of Georgia; MIT; Max Planck Institute)}  & $1,474$ & $2.4$M images&\checkmark&\checkmark&$\times$	& Collected by mobile devices in daily life; Variable lighting condition and head motion. &\small{\url{https://gazecapture.csail.mit.edu}}\\
				\hline
				MPIIFaceGaze~\cite{Zhang_2017_CVPRW}, \textcolor{white}{***} 2017,  {\small(Max Planck Institute)} 	  & $15$ & $\sim45$K images&\checkmark & \checkmark &\checkmark  & Collected by laptops in daily life; Free head pose and illumination. &{\small \color[RGB]{236,0,140}footnote\tnote{1}}\\
				\hline
				InvisibleEye~\cite{Tonsen_2017_IMWUT}, 2017, {\small(Max Planck Institute; Osaka University)}&17&280K Images &$\times$&\checkmark&$\times$&Collected in laboratory; Multiple near-eye camera; Low resolution cameras .&\small{\url{https://mpi-inf.mpg.de/invisibleeye}}\\
				\hline
				TabletGaze~\cite{Huang_2017_mva}, 2017,\textcolor{white}{*} {\small(Rice University)}	& $51$	&$816$ videos	&\checkmark&\checkmark&$\times$&Collected by tablets in laboratory; Four postures to hold the tablets; Free head pose. &\small{\url{https://sh.rice.edu/cognitive-engagement/ tabletgaze}}\\
				\hline
				RT-Gene~\cite{Fischer_2018_ECCV}, 2018,\textcolor{white}{****}  {\small (Imperial College London)}	& $15$	& $123$K images &\checkmark&$\times$&\checkmark&Collected in laboratory; Free head pose; Annotated with mobile eye-tracker; Use GAN to remove the eye-tracker in face images. &\small{\url{https://github.com/Tobias-Fischer/rt_gene}} \\ 
				\hline
				Gaze360~\cite{Kellnhofer_2019_ICCV}, 2019, \textcolor{white}{****} {\small(MIT; Toyota Research Institute)} & $238$ & $172$K images &\checkmark&$\times$&\checkmark &Collected in indoor and outdoor environments; A wide range of head poses and distances between subjects and cameras. & \small{\url{https://gaze360.csail.mit.edu}} \\  
				\hline
				NVGaze~\cite{Kim_2019_CHI}, 2019, \textcolor{white}{***} {\small(NVIDIA; UNC)}&30 & 4.5M images  &$\times$&\checkmark&$\times$&Collected in laboratory; Near-eye Images; Infrared illumination. &\small{\url{https://sites.google.com/nvidia.com/nvgaze}}\\
				\hline
				ShanghaiTechGaze~\cite{Lian_2019_TNNLS},\textcolor{white}{*} 2019, {\small(ShanghaiTech University; UESTC)}&$137$ & $224$K images&\checkmark&\checkmark&$\times$  &Collected in laboratory; Free head poes; Multiview gaze dataset.& \small{\url{https://github.com/dongzelian/multi-view-gaze}} \\
				\hline
				ETH-XGaze~\cite{Zhang_2020_ECCV}, 2020,  \textcolor{white}{*} {\small(ETH Zurich; Google)}&110&1.1M images &\checkmark&\checkmark&\checkmark&Collected in laboratory; High-resolution images; Extreme head pose; 16 illumination conditions.&\small{\url{https://ait.ethz.ch/projects/2020/ETH-XGaze}}\\ 
				\hline
				EVE~\cite{park_2020_eccv}, 2020,\textcolor{white}{******}  {\small(ETH Zurich)} &54&$\sim4.2$K videos&\checkmark&\checkmark&\checkmark&Collected in laboratory; Free head pose; Free view; Annotated with desktop eye tracker; Pupil size annotation.&\small{\url{https://ait.ethz.ch/projects/2020/EVE/}}\\
				\bottomrule
			\end{tabular}
			\begin{tablenotes}
				\footnotesize
				\item[1] \url{https://mpi-inf.mpg.de/departments/computer-vision-and-machine-learning/research/gaze-based-human-computer-interaction/its-written-all-over-your-face-full-face-appearance-based-gaze-estimation}
			\end{tablenotes}
		\end{threeparttable}
		
		\label{table:statistics}
	\end{table*}
	
	\begin{table*}[!htp]
		\setlength\tabcolsep{8pt}
		\renewcommand\arraystretch{1.2}
		\renewcommand{\multirowsetup}{\centering}
		\caption{\added{Benchmark of within-dataset evaluation. We use the provided source codes or re-implement ($^{\dag}$) the methods for comparison. The \underline{underlines} indicate the top three best performances. Note that the methods in the last row are proposed for point of gaze estimation, we convert the result using the post-processing method in \sref{ssec_datapost}.}}
		\vspace{-3mm}
		\begin{threeparttable}
			\begin{tabular}
            {c|c|ccc|ccccc}
				\toprule
				\textbf{Methods}&Pub.&\makecell{MPIIGaze\\\cite{Zhang_2017_tpami}}&\makecell{EyeDiap\\\cite{Mora_2014_ETRA}} &\multicolumn{1}{c|}{\makecell{UT\\\cite{Sugano_2014_CVPR}}}&\makecell{MPIIFaceGaze\\\cite{Zhang_2017_CVPRW}}&\makecell{EyeDiap\\\cite{Mora_2014_ETRA}}&\makecell{Gaze360\\\cite{Kellnhofer_2019_ICCV}}&\makecell{RT-Gene\\\cite{Fischer_2018_ECCV}}&\makecell{ETH-XGaze\\\cite{Zhang_2020_ECCV}}\\
				
    		\hline
                \hline
                Mnist$^{\dag}$~\cite{Zhang_2015_CVPR}&CVPR15&$6.27^\circ$&$7.60^\circ$&\first{$6.34^\circ$}&$6.39^\circ$&$7.37^\circ$&N/A&N/A&N/A\\
				GazeNet$^{\dag}$~\cite{Zhang_2017_tpami}&TPAMI17&$5.70^\circ$&$7.13^\circ$&\second{$6.44^\circ$}&$5.76^\circ$&$6.79^\circ$&N/A&N/A&N/A\\
			
				\hline
				Dilated-Net$^{\dag}$~\cite{Chen_2019_ACCV}&ACCV19&$4.39^\circ$&$6.57^\circ$&N/A&$4.42^\circ$&$6.19^\circ$&$13.73^\circ$&$8.38^\circ$&N/A\\
				Gaze360~\cite{Kellnhofer_2019_ICCV}&ICCV19&\first{$4.07^\circ$}&\first{$5.58^\circ$}&N/A&\second{$4.06^\circ$}&$5.36^\circ$&\second{$11.04^\circ$}&\second{$7.06^\circ$}&\first{$4.46^\circ$}\\
				RT-Gene~\cite{Fischer_2018_ECCV}&ECCV18&$4.61^\circ$&\third{$6.30^\circ$}&N/A&$4.66^\circ$&$6.02^\circ$&$12.26^\circ$&$8.60^\circ$&N/A\\
				FullFace~\cite{Zhang_2017_CVPRW}&CVPRW17&$4.96^\circ$&$6.76^\circ$&N/A&$4.93^\circ$&$6.53^\circ$&$14.99^\circ$&$10.00^\circ$&\second{$7.38^\circ$}\\
			RCNN$^{\dag}$~\cite{Palmero_2018_BMVC}& BMVC18 & N/A&N/A&N/A&$4.10^\circ$&\third{$5.31^\circ$}&$11.23^\circ$&$10.30^\circ$&N/A\\
                CA-Net~\cite{Cheng_2020_AAAI}&AAAI20&\second{$4.27^\circ$}&\second{$5.63^\circ$}&N/A&$4.27^\circ$&\second{$5.27^\circ$}&\third{$11.20^\circ$}&\third{$8.27^\circ$}&N/A\\
                GazeTR-Pure~\cite{cheng2021gaze}&ICPR22&N/A&N/A&N/A&$4.74^\circ$&$5.72^\circ$&$13.58^\circ$&$8.06^\circ$&N/A\\
                GazeTR-Hybird~\cite{cheng2021gaze}&ICPR22& N/A&N/A&N/A&\first{$4.00^\circ$}&\first{$5.17^\circ$}&\first{$10.62^\circ$}&\first{$6.55^\circ$}&N/A\\
				\hline

				Itracker$^{\dag}$~\cite{Krafka_2016_CVPR}&CVPR16&$7.25^\circ$&$7.50^\circ$&N/A&$7.33^\circ$&$7.13^\circ$&N/A&N/A&N/A\\
				AFF-Net~\cite{Bao_2020_ICPR}&ICPR20&\third{$3.69^\circ$}&$6.75^\circ$&N/A&\third{$3.73^\circ$}&$6.41^\circ$&N/A&N/A&N/A\\
				\bottomrule
				
			\end{tabular}

		\end{threeparttable}
            \vspace{-3mm}
		\label{tab:3dgaze}
	\end{table*}

	To derive the 3D gaze target $\bm{t}$, we obtain the pose $\{\bm{R_s}, \bm{T_s}\}$ of screen coordinate system (SCS)~\wrt~CCS by geometric calibration, where $\bm{R_s}$ is the rotation matrix and $\bm{T_s}$ is the translation matrix. 
	The $\bm{t}$ is computed as $\bm{t} = \bm{R_s}[u, v, 0]^T+\bm{T_s}$, where the additional $0$ is the $z$-axis coordinate of $\bm{p}$ in SCS. 
	The 3D gaze origin $\bm{o}$ is usually defined as the face center or the eye center. 
	It can be estimated by landmark detection algorithms or stereo measurement methods.
	
	On the other hand, given a 3D gaze direction $\bm{g}$, we aim to compute the corresponding 2D target point $\bm{p}$ on the screen.	
	Note that, we also need to acquire the screen pose $\{\bm{R_s}, \bm{T_s}\}$ as well as the origin point $\bm{o}$ as mentioned previously. We first compute the intersection of gaze direction and screen,~\ie, 3D gaze target $\bm{t}$, in CCS, and then we convert the 3D gaze target to the 2D gaze target using the pose $\{\bm{R_s}, \bm{T_s}\}$.
	
	To deduce the equation of screen plane, we compute $\bm{n} = \bm{R_s}[:,2] = (n_x, n_y, n_z)$, where $\bm{n}$ is the normal vector of screen plane.
	$\bm{T_s}=[t_x, t_y, t_z]^T$ also represents a point on the screen plane. Therefore, the equation of the screen plane is 
	\begin{equation}
		\label{equ:screenplane}
		n_xx + n_yy + n_zz = n_xt_x + n_yt_y + n_zt_z.
	\end{equation}
	
	Given a gaze direction $\bm{g}$ and the origin point $\bm{o}$, we can write the equation of the line of sight as
	\begin{equation}
		\label{equ:gazeline}
		\frac{x - x_o}{g_x} = \frac{y - y_o}{g_y} =\frac{z - z_o}{g_z}.
	\end{equation}
	
	By solving \eref{equ:screenplane} and \eref{equ:gazeline}, we obtain the intersection $\bm{t}$, and $(u, v, z) = \bm{R_s}^{-1}(\bm{t}-\bm{T_s})$, where $z$ usually equals to $0$ and $\bm{p}=(u, v)$ is the coordinate of 2D target point in metre.

 	\begin{figure*}[t]
		\centering
		\includegraphics[width=\columnwidth*2]{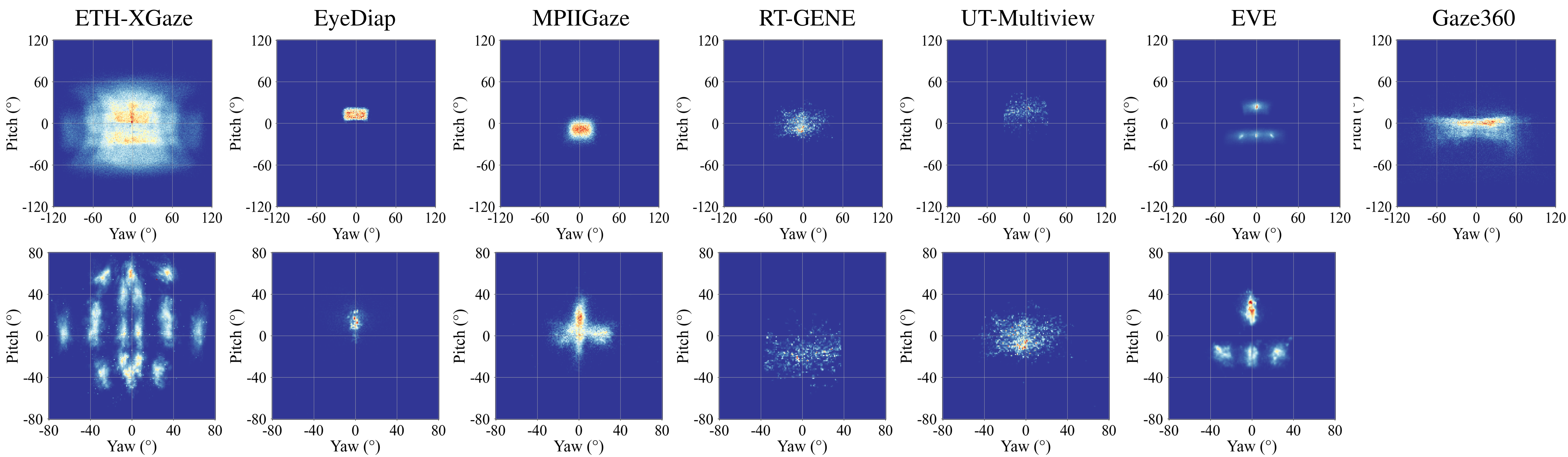} 
		\vspace{-3mm}
		\caption{Distribution of head pose and gaze in different datasets. The first row is the distribution of gaze and the second row show head distribution.}
        \vspace{-5mm}
		\label{fig:datasetrange}
	\end{figure*}
	
	\subsubsection{Gaze Origin Conversion}
	\label{sssec_vector}
	Conventional gaze estimation methods usually estimate gaze directions \wrt~each eye. 
	They define the origin of gaze directions as each eye center~\cite{Zhang_2017_tpami, Park_2018_ECCV, Wang2_2019_CVPR, Liu_2019_tpami}. Recently, more attention has been paid to gaze estimation using the face images and they estimate gaze direction \wrt the whole face. They define the gaze vector starting from the face center to the gaze target~\cite{Cheng_2020_AAAI, Park_2019_ICCV, Zhang_2017_CVPRW, Fischer_2018_ECCV}.	
	Here we introduce a gaze origin conversion method to bridge the gap between these two types of gaze estimates.
	
	We first compute the pose $\{\bm{R_s}, \bm{T_s} \}$ of SCS and the origin $\bm{o}$ of the predicted gaze direction $\bm{g}$ through calibration. 
	Then we can write \eref{equ:screenplane} and \eref{equ:gazeline} based on these parameters. 
	The 3D gaze target point $\bm{t}$ can be calculated by solving the equation of \eref{equ:screenplane} and \eref{equ:gazeline}. 
	Next, we obtain the new origin $\bm{o_n}$ of the gaze direction through 3D landmark detection. The new gaze direction can be computed by 
	\begin{equation}
		\bm{g_{new}} = \frac{t-o_n}{||t-o_n||}.
	\end{equation}

	\subsection{Evaluation Metrics}
	\label{ssec_metric}

	Two types of metric are used for performance evaluation: angular error and the Euclidean distance. 
	The angular error measures the accuracy of 3D gaze estimation methods ~\cite{Cheng_2020_AAAI,Park_2019_ICCV,Zhang_2017_tpami}. Assuming the actual gaze direction is $\bm{g}\in\mathbb{R}^3$ and the estimated gaze direction is $\bm{\hat{g}}\in\mathbb{R}^3$, the angular error can be computed as:
	\begin{equation}
		\label{equ:3dmetric}
		\mathcal{L}_{\mathrm{angular}} = \frac{\bm{g}\cdot\bm{\hat{g}}}{||\bm{g}||||\bm{\hat{g}}||}
	\end{equation}
	
\noindent	The Euclidean distance has been used for measuring the accuracy of 2D gaze estimation methods in~\cite{Krafka_2016_CVPR,Chang_2019_iccvw,Wong_2019_Percom}. We denote the actual gaze position as $\bm{p}\in\mathbb{R}^2$ and the estimated gaze position as $\bm{\hat{p}}\in\mathbb{R}^2$. We can compute the Euclidean distance as
	\begin{equation}
		\label{equ:2dmetric}
		\mathcal{L}_{\mathrm{Euclidean}} = ||\bm{p} - \bm{\hat{p}}||_2,
	\end{equation}


    Two kinds of evaluation protocols are commonly used for deep-learning based gaze estimation methods, including within-dataset and cross-dataset evaluation. 
	The within-dataset evaluation assesses the model performance on the unseen subjects from the same dataset. The dataset is divided into training and test set according to subjects. There is no overlap in subjects between the training and test set. Note that, most of the gaze datasets provide within-dataset evaluation protocol. They divide the data into training and test set in advance. 
	The cross-dataset evaluation assesses the model performance on the unseen environment. The model is trained on one dataset and tested on another dataset.

	\begin{table*}[htp]
		\setlength\tabcolsep{9pt}
		\centering
		\renewcommand\arraystretch{1.2}
		\renewcommand{\multirowsetup}{\centering}
		\caption{\added{ Benchmark of cross-domain gaze estimation. `Source-free' indicates whether the method requires source images during domain adaption. `Target num' presents the number of images used for domain adaption. $\mathcal{D}_E$, $\mathcal{D}_G$, $\mathcal{D}_M$, $\mathcal{D}_D$ denotes ETH-XGaze~\cite{Zhang_2020_ECCV}, Gaze360~\cite{Kellnhofer_2019_ICCV}, MPIIGaze~\cite{Zhang_2017_CVPRW} and EyeDiap~\cite{Mora_2014_ETRA} datasets. The second-row methods use unannotated images while the third-row methods use annotated images.}}
		\vspace{-3mm}
			\begin{tabular}{l|cc|cc|cccc}
				\toprule
				Methods & Pub. & Year &Source-free & Target num&$\mathcal{D}_E\rightarrow\mathcal{D}_M$ & $\mathcal{D}_E\rightarrow\mathcal{D}_D$ & $\mathcal{D}_G\rightarrow\mathcal{D}_M$ & $\mathcal{D}_G\rightarrow\mathcal{D}_D$\\
				\hline
				\hline
				FullFace~\cite{Zhang_2017_CVPRW} & CVPRW & 2017 & \multirow{4}{*}{\makecell{\textit{w/o} \\adaption}}& \multirow{4}{*}{\makecell{\textit{w/o} \\adaption}}& $11.13^\circ$ & $14.42^\circ$ &$12.35^\circ $&$30.15^\circ$\\
				CA-Net~\cite{Cheng_2020_AAAI} & AAAI & 2020 & &  & \na& \na& $27.13^\circ$ & $31.41^\circ$ \\
				
				PureGaze~\cite{cheng_2022_aaai} & AAAI &2022 & & &$9.28^\circ$& $9.32^\circ$& $7.08^\circ$& $7.48^\circ$\\
				RAT~\cite{Bao_2022_CVPR} & CVPR &2022 & & &$7.40^\circ$&$6.91^\circ$&$7.69^\circ$&$7.08^\circ$\\
				\hline
				PnP-GA\cite{liu2021generalizing} & ICCV &2021 & $\times$ & 10 &$6.00^\circ$&$6.17^\circ$&\underline{$5.74^\circ$}&$7.04^\circ$\\
				CSA\cite{wang2022contrastive}& CVPR &2022&$\checkmark$ & unreport &\underline{$5.37^\circ$} &$6.77^\circ$ &$7.30^\circ$ &$7.73^\circ$\\
				CRGA-100\cite{wang2022contrastive}&CVPR &2022&$\times$&100 &$5.68^\circ$ & \third{$5.72^\circ$} & $6.09^\circ$&\third{$6.68^\circ$} \\
				CRGA~\cite{wang2022contrastive}&CVPR& 2022&$\times$&unreport &\third{$5.48^\circ$} & \second{$5.66^\circ$} & \third{$5.89^\circ$}&\second{$6.49^\circ$}\\
				RUDA~\cite{Bao_2022_CVPR} &CVPR &2022&$\times$& 100 & $5.78^\circ$ & \underline{$5.10^\circ$}&$6.88^\circ$&$6.73^\circ$\\
				\hline
				\makecell[l]{PureGaze-FT~\cite{cheng_2022_aaai}} & AAAI &2022 &$\checkmark$ &$\sim50$ &\first{$5.20^\circ$}& $7.36^\circ$& \first{$5.30^\circ$}& \first{$6.42^\circ$}\\
				\bottomrule
						
				
			\end{tabular}
		\vspace{-5mm}
		\label{tab:crossdataset}
	\end{table*}
	
	\begin{table}[t]
		\setlength\tabcolsep{4pt}
		\centering
		\renewcommand\arraystretch{1.2}
		\renewcommand{\multirowsetup}{\centering}
		\caption{Benchmark of 2D gaze estimation (cm).}
		\vspace{-3mm}
		\begin{threeparttable}
			\begin{tabular}{l|>{\fontsize{9}{12}\selectfont}ccccc}
				\toprule
				\multirow{2}{*}{Methods} &\multirow{2}{*}{\normalsize Pub.}&\multirow{2}{*}{\makecell{MPIIGaze\\~\cite{Zhang_2017_CVPRW}}} & \multirow{2}{*}{\makecell{EyeDiap\\~\cite{Mora_2014_ETRA}}}&\multicolumn{2}{c}{GazeCapture~\cite{Krafka_2016_CVPR}}\\
				&&&&Tablet&Phone\\
				
				\hline
				\hline
				Itracker$^\dag$~\cite{Krafka_2016_CVPR}&CVPR16&7.67&10.13 &2.81&1.86 \\
				AFF-Net~\cite{Bao_2020_ICPR}&ICPR20&\first{4.21}&9.25 &\first{2.30} &\second{1.62} \\
				SAGE~\cite{He_2019_iccvw}&ICCVW19&N/A&N/A&{2.72} &{1.78} \\
				TAT~\cite{Guo_2019_iccvw}&ICCVW19&N/A&N/A&\third{2.66} &\third{1.77}\\
				\added{EFE~\cite{Balim_2023_CVPR}}& \added{CVPRW23}&\added{\first{3.89}} & \added{N/A} & \added{\second{2.48}} & \added{\first{1.61}}\\
				\hline
				Mnist$^\dag$~\cite{Zhang_2015_CVPR}&CVPR15&7.29&9.06 &N/A&N/A\\
				GazeNet$^\dag$~\cite{Zhang_2017_tpami}&TPAMI17&6.62 &8.51 &N/A&N/A\\
				\hline
			DilatedNet$^\dag$~\cite{Chen_2019_ACCV}&ACCV19&5.07 &7.36 &N/A&N/A\\
			Gaze360~\cite{Kellnhofer_2019_ICCV}&ICCV19&\second{4.66} &\second{6.37} &N/A&N/A\\
			RT-Gene~\cite{Fischer_2018_ECCV}&ECCV18&5.36  &\second{7.19}  &N/A&N/A\\
			FullFace~\cite{Zhang_2017_CVPRW}&CVPRW17&5.65 &7.70 &N/A&N/A\\
			CA-Net~\cite{Cheng_2020_AAAI}&AAAI20&4.90 &\first{6.30} &N/A&N/A\\
			\bottomrule
		\end{tabular}
	\end{threeparttable}
	\vspace{-5mm}
	\label{tab:2dgaze}
\end{table}

	\subsection{Public Datasets}
	\label{ssec_dataset}
	\added{We try our best to summarize all the public datasets on gaze estimation, as shown in \Tref{table:statistics}. The gaze and head pose distribution of these datasets are shown in~\fref{fig:datasetrange}.} Note that, the Gaze360 dataset do not provide the head information.
	We also discuss three typical datasets that are widely used in gaze estimation studies.

	Zhang \etal proposed the MPIIGaze~\cite{Zhang_2015_CVPR} dataset.
	It is the most popular dataset for appearance-based gaze estimation methods. It contains a total of 213,659 images collected from 15 subjects. The images are collected in daily life over several months and there is no constraint for the head pose. MPIIGaze dataset provides both 2D and 3D gaze annotation. It also provides a standard evaluation set, which contains 15 subjects and 3,000 images for each subject. 
	The 3000 images are consisted of 1,500 left-eye images and 1,500 right-eye images. 
	The author further extends the original datasets in ~\cite{Zhang_2017_CVPRW,Zhang_2017_tpami}. They supply the corresponding face images in \cite{Zhang_2017_CVPRW} and manual landmark annotations in \cite{Zhang_2017_tpami}. 
	
	
	EyeDiap~\cite{Mora_2014_ETRA} dataset consists of 94 video clips from 16 participants. \added{It is collected in laboratory environments and has three visual target sessions:  continuous moving targets, discrete moving targets, and floating ball.} For each subject, they recorded a total of six sessions containing two head movements: static head pose and free head movement. Two cameras are used for data collection: an RGBD camera and an HD camera. \added{The disadvantage of this dataset is the lack of illumination variation.}
	

	GazeCapture~\cite{Krafka_2016_CVPR} dataset is collected through crowdsourcing. It contains a total of 2,445,504 images from 1,474 participants. All images are collected using mobile phones or tablets. Each participant is required to gaze at a circle shown on the devices without any constraint on their head movement. As a result, the GazeCapture dataset covers various lighting conditions and head motions. The GazeCapture dataset does not provide 3D coordinates of targets. It is usually used for the evaluation of unconstrained 2D gaze point estimation methods.
	

	In addition to the dataset mentioned above, there are several datasets being proposed recently. 
	For example, in 2018, Fischer~\etal proposed RT-Gene dataset~\cite{Fischer_2018_ECCV}. This dataset provides accurate 3D gaze data since they collect gaze with a dedicated eye tracking device. In 2019, Kellnhofe~\etal proposed the Gaze360 dataset~\cite{Kellnhofer_2019_ICCV}. The dataset consists of 238 subjects of indoor and outdoor environments with 3D gaze across a wide range of head poses and distances. In 2020, Zhang~\etal propose the ETH-XGaze dataset~\cite{Zhang_2020_ECCV}. This dataset provides high-resolution images that cover extreme head poses.
    It also contains 16 illumination conditions for exploring the effects of illumination.
	
    \subsection{Benchmarks}
    \added{We build benchmarks for 2D PoG and 3D gaze estimation in this section. 
    We re-implemented the typical gaze estimation methods as annotated with $^{\dag}$ or report the performance from their manuscripts for comparison.}
    Note that, 2D PoG estimation methods and 3D gaze estimation methods are not comparable since they estimate different forms of gaze.
    We follow \sref{sssec_2d3d} to convert the estimation results.
    We convert 2D PoG into 3D gaze and vice versa. 
    Besides, there are two different gaze definitions in 3D gaze estimation methods.
    Conventional methods define the origin of gaze direction as eye centers. Recent methods estimate gaze from face images where the gaze origin is defined as face centers.
    The difference between the two definitions is minor but makes the direct comparison unfair.
    We also convert the two definitions with post-processing methods following \sref{sssec_vector}.
    We respectively conduct benchmarks for 2D PoG and 3D gaze estimation. The 3D gaze estimation also are divided into within-dataset and cross-dataset evaluation. We mark the top three performance in all benchmarks with underlines.

    \noindent \textbf{Within-dataset evaluation.} 
    We first show the comparison of within-dataset evaluation in~\Tref{tab:3dgaze}. 
    The second row contains methods estimating 3D gaze from eye images where the gaze origin is eye center.
    The methods in the third row estimate 3D gaze from face images. They define the gaze origin as face centers.
    The last row contains the methods which estimate 2D PoG from face images.
    \added{Evaluation datasets contain two categaries based on data pre-processing process.
    We obtain eye images from MPIIGaze~\cite{Zhang_2017_tpami}, EyeDiap~\cite{Mora_2014_ETRA} and UT~\cite{Sugano_2014_CVPR}, and evaluate the method which define eye centers as gaze origin in the three datasets.}
    The result is shown in the third column of ~\Tref{tab:3dgaze}.
    We obtain face and eye images from MPIIFaceGaze~\cite{Zhang_2017_CVPRW}, EyeDiap~\cite{Mora_2014_ETRA}, Gaze360~\cite{Kellnhofer_2019_ICCV}, RT-Gene~\cite{Fischer_2018_ECCV} and ETH-XGaze datasets~\cite{Zhang_2020_ECCV}.
    We evaluate the method which defines face centers as gaze origin in these datasets.
    The result is shown in the fourth column of ~\Tref{tab:3dgaze}.
    
 	\added{Conventional approaches typically estimate gaze using eye images. The Mnist~\cite{Zhang_2015_CVPR} and GazeNet~\cite{Zhang_2017_tpami} methods employ eye images and head pose vector as input for gaze estimation.
 	Recent methods, \ie, the third-row methods, focus on estimating gaze from facial images.
 	Despite incurring higher computational costs, methods reliant on facial images outperform those centered on eye images. 
 	Notably, face image-based methods also usually maintain an acceptable inference speed exceeding 20 frames per second.}
 	
 	\added{Among face image-based methods, GazeTR-Hybird~\cite{cheng2021gaze}, CA-Net~\cite{Cheng_2020_AAAI} and Gaze360~\cite{Kellnhofer_2019_ICCV} have better performance. Gaze360 employs ResNet18 for feature extraction while GazeTR-Hybrid adopts a mixed architecture of ResNet18 and transformers. 
 	Pre-training significantly enhances the performance of the two methods. 
 	In contrast, CA-Net leverages features from both facial and eye images,
 	It requires no pre-training but has a complex network.}
 	\added{Regarding datasets, Gaze360~\cite{Kellnhofer_2019_ICCV} and RT-Gene~\cite{Fischer_2018_ECCV} are collected with large user-camera distances. 
 	Most of methods demonstrate significant errors in the two datasets due to low-resolution images.
 	Other datasets are collected with a small user-camera distance or with high-resolution cameras. Appearance-based gaze estimation methods usually achieve approximately $5^\circ$ in these environments.} 
          
    \noindent \textbf{Cross-Dataset Evaluation.} We conduct four tasks including $\mathcal{D}_E\rightarrow\mathcal{D}_M$, $\mathcal{D}_E\rightarrow\mathcal{D}_D$, $\mathcal{D}_G\rightarrow\mathcal{D}_M$ and $\mathcal{D}_G\rightarrow\mathcal{D}_D$, where $\mathcal{D}_E$, $\mathcal{D}_G$, $\mathcal{D}_M$ and $\mathcal{D}_D$ represents ETH-XGaze~\cite{Zhang_2020_ECCV}, Gaze360~\cite{Kellnhofer_2019_ICCV}, MPIIFaceGaze~\cite{Zhang_2017_CVPRW} and EyeDiap~\cite{Mora_2014_ETRA} datasets.
    \added{ETH-XGaze and Gaze360 are used as training set since they have large gaze and head pose ranges.}
    The result is shown in~\Tref{tab:crossdataset}. \added{Unsupervised domain adaption methods are usually proposed to solve the cross-dataset problem. These methods require target images for domain adaption. We summarize the number of required target images. The source-free column indicates whether the method requires source images during domain adaption.}

    The methods in the second row train models on source datasets without adaption.
    \added{PureGaze~\cite{cheng_2022_aaai} and RAT~\cite{Bao_2022_CVPR} integrate specific algorithms to enhance model generalization.
    Their models can be directly applied into multiple domains and achieve reasonable performance.
    The third row shows the performance of unsupervised domain adaption methods.}
    CRGA~\cite{wang2022contrastive} and RUDA~\cite{Bao_2022_CVPR} have better performance while PnP-GA~\cite{liu2021generalizing} has lower requirement.
    \added{Compared with the second-row methods, these methods leverage target images to improve the model performance within specific domains. This approach yields a dedicated model for each domain, outperforming PureGaze and RAT.
    Notably, CSA~\cite{wang2022contrastive} stands out as a source-free method that dispenses with the need for a source dataset during adaptation. 
    This trend is noteworthy for its implications in privacy protection.}
    PureGaze-FT~\cite{cheng_2022_aaai} samples 5 images per person for fine-tuning. Although the method achieves good performance with $50$ images, it requires annotated images while previous methods only require unannotated images.
    
    \textbf{2D PoG estimation.} We conduct experiment for 2D PoG estimation. 
    We use MPIIGaze~\cite{Zhang_2017_tpami}, EyeDiap~\cite{Mora_2014_ETRA} and GazeCapture~\cite{Krafka_2016_CVPR} for evaluation sets and Euclidean distance for evaluation metric. MPIIGaze and EyeDiap datasets collect 2D PoG in screen. The two datasets both provide calibrated screen pose, where we can convert gaze directions to 2D PoG. GazeCapture dataset collects 2D PoG in mobile devices. We count the result based on the types of devices, \eg, tablets and phones.
    The second row in \Tref{tab:2dgaze} shows the result of PoG estimation methods. \added{AFF-Net~\cite{Bao_2020_ICPR} and EFE~\cite{Balim_2023_CVPR} shows the best performance than other compared methods.}
    The third and fourth rows show the converted results. Compared methods are designed for gaze direction estimation and we convert the result into PoG.
    The converted result shows good accuracy in EyeDiap dataset while AFF-Net also shows the best performance in MPIIGaze dataset.
    \vspace{-1mm}

	\section{Conclusions and Future Directions}
	\label{sec_conclusion}
	
	In this survey, we present a comprehensive overview of deep learning-based gaze estimation methods. Unlike the conventional gaze estimation methods that requires dedicated devices, the deep learning-based approaches regress the gaze from the eye appearance captured by web cameras. This makes it easy to implement the algorithm in real world applications.
	We introduce the gaze estimation method from four perspectives: deep feature extraction, deep neural network architecture design, personal calibration as well as device and platform.
	We summarize the public datasets on appearance-based gaze estimation and provide benchmarks to compare of the state-of-the-art algorithms. 
	This survey can serve as a guideline for future gaze estimation research.

	We further suggest several future directions of deep learning-based gaze estimation.
	1) Extracting more robust gaze features. The perfect gaze estimation method should be accurate under all different subjects, head poses and environments. Therefore, an environment-invariant gaze feature is crucial.
	2) Improving performance with fast and simple calibration. There is a trade-off between the system performance and calibration time. The longer calibration time leads to more accurate estimates. How to achieve satisfactory performance with fast calibration procedure is a promising direction. 
	3) Interpreting learned features. Deep learning approaches often serve as a black-box tool for gaze estimation. Interpretation of the learned features in these methods will bring insight for the deep learning-based gaze estimation.

    \vspace{-3.5mm}

	\footnotesize
	\bibliographystyle{IEEEtran}
	\bibliography{gaze}

	\begin{IEEEbiography}[{\includegraphics[width=0.9in,height=1.25in,clip,keepaspectratio]{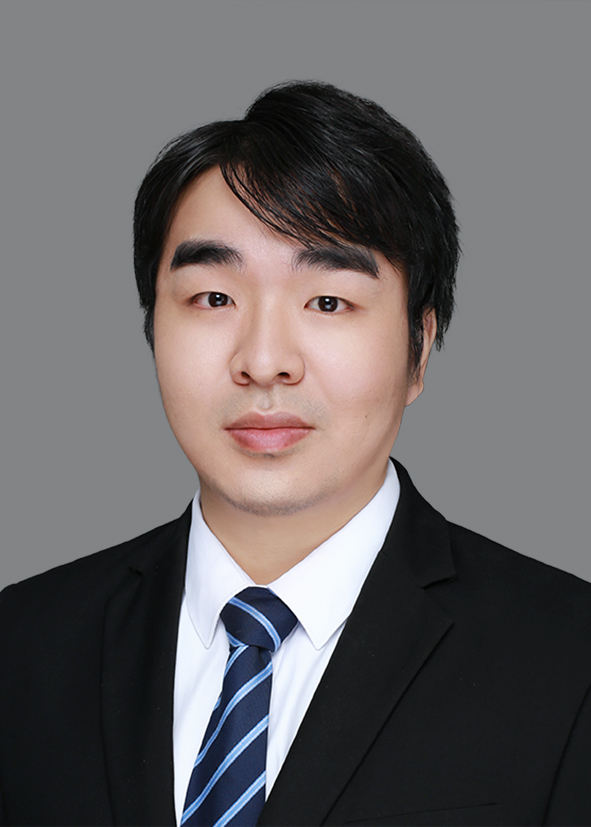}}]{Yihua Cheng} received the B.S. degree in computer science from Beijing University of Posts and Telecommunications in 2017, and Ph.D. degree in computer science from Beihang University in 2022. He is now a Postdoctoral Researcher with University of Birmingham, UK. His research interests include gaze estimation, hand pose estimation, object pose estimation and human-robot interaction.
	\end{IEEEbiography}
	
	\begin{IEEEbiography}[{\includegraphics[width=1in,height=1.25in,clip,keepaspectratio]{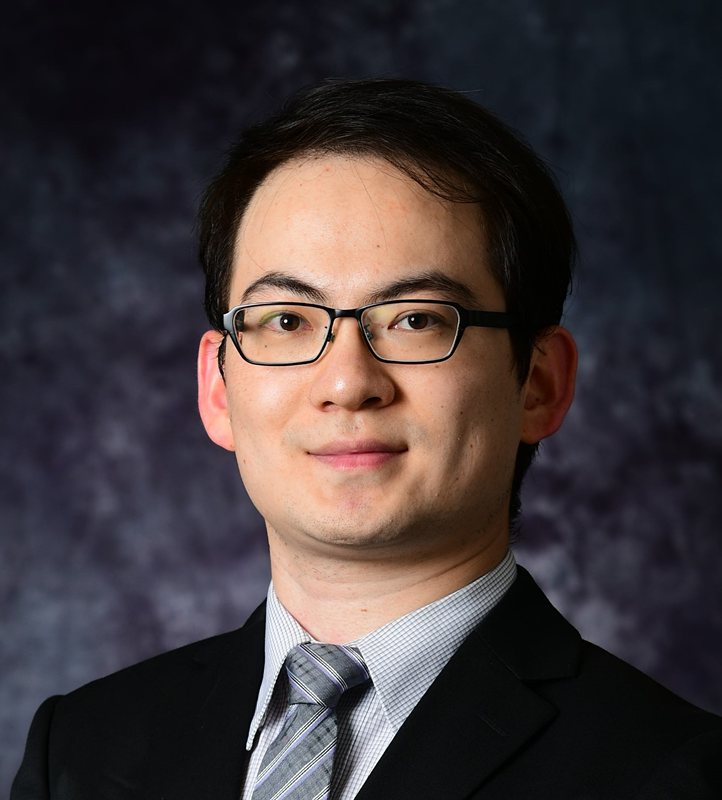}}]{Haofei Wang}
		received the B.S. degree with distinction from Zhejiang University in 2013, and Ph.D. degree in electronic and computer engineering from the Hong Kong University of Science and Technology in 2020. He is now a Postdoctoral Researcher with the Pengcheng Laboratory, Shenzhen, China. His research interests include eye tracking, gaze estimation, human-computer interaction and mixed reality.
	\end{IEEEbiography}
	
	\begin{IEEEbiography}[{\includegraphics[width=1in,height=1.25in,clip,keepaspectratio]{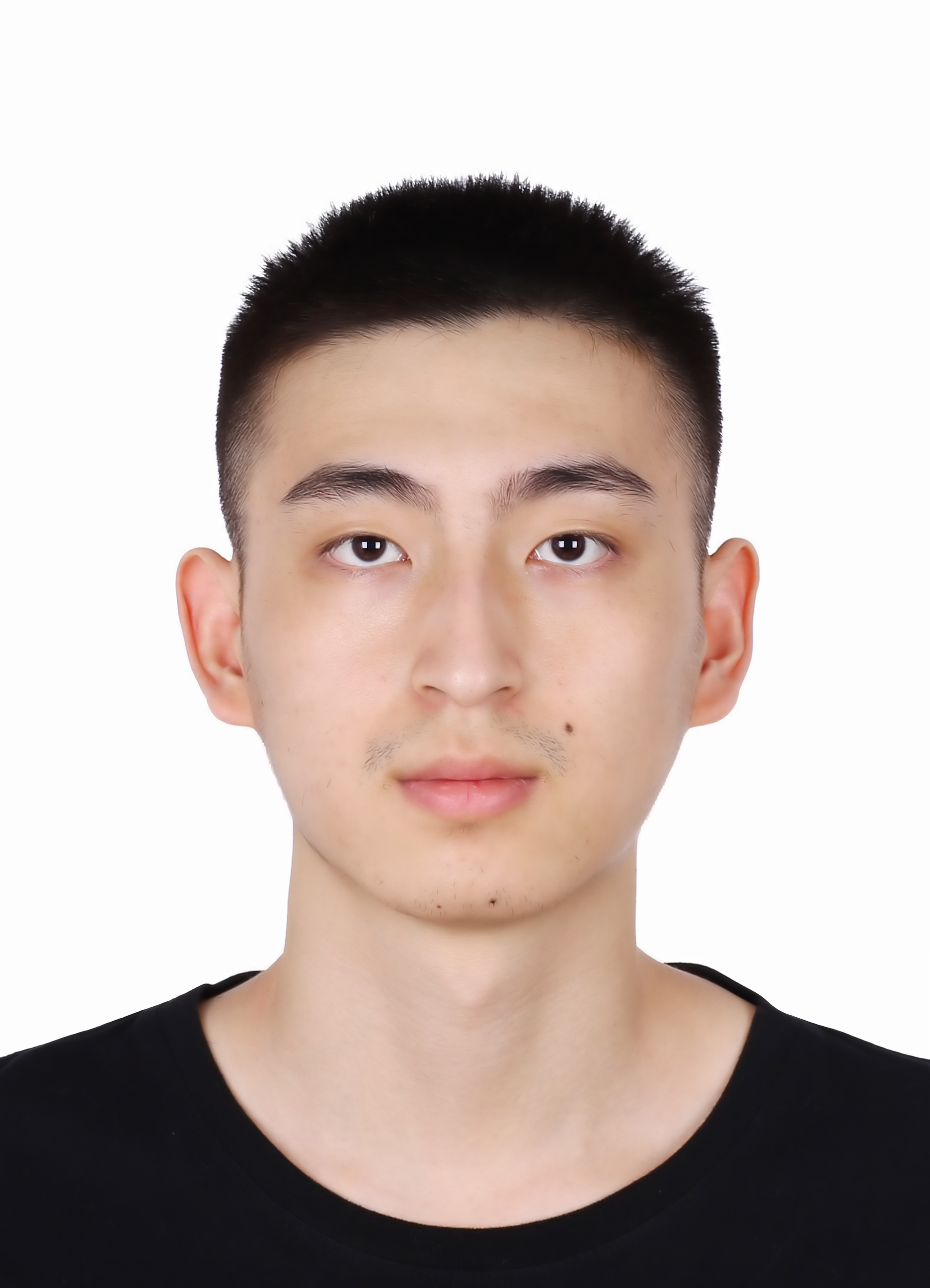}}]{Yiwei Bao}
		currently pursuing the Ph.D. degree with the State Key Laboratory of Virtual Reality Technology and Systems, School of Computer Science and Engineering, Beihang University. His research interests include computer vision and human gaze analysis.
	\end{IEEEbiography}
	
	\begin{IEEEbiography}[{\includegraphics[width=1in,height=1.25in,clip,keepaspectratio]{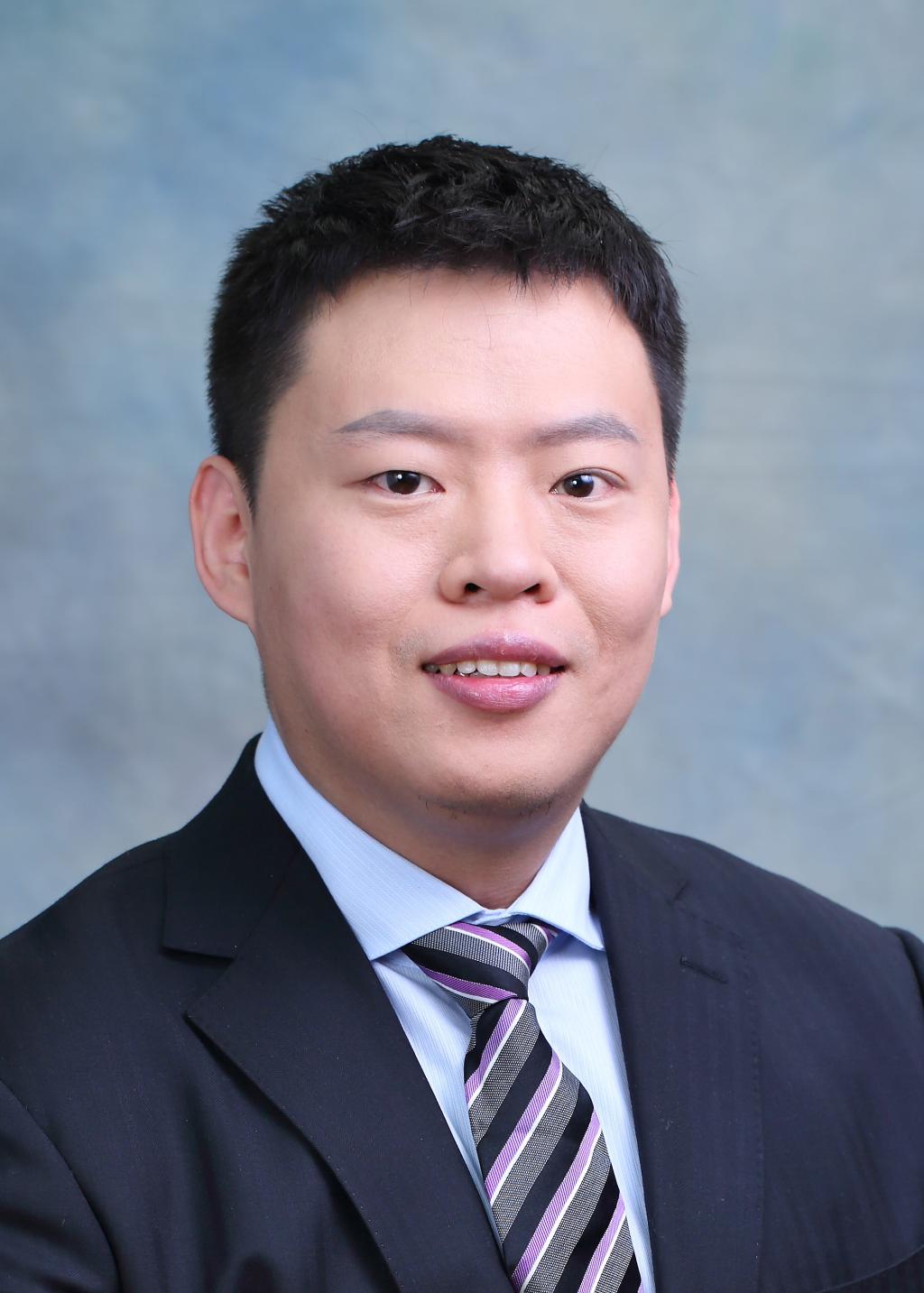}}]{Feng Lu}
		received the B.S. and M.S. degrees in automation from Tsinghua University, in 2007 and 2010, respectively, and the Ph.D. degree in information science and technology from The University of Tokyo, in 2013. He is currently a Professor with the State Key Laboratory of Virtual Reality Technology and Systems, School of Computer Science and Engineering, Beihang University. His research interests include computer vision, human-computer interaction and augmented intelligence.
	\end{IEEEbiography}
	

	
	

\end{document}